\documentclass[sn-mathphys-num,iicol]{sn-jnl}% Default with double column layout
\usepackage[table,dvipsnames,xcdraw]{xcolor}
\usepackage{colortbl}
\usepackage{graphicx}%
\usepackage{multirow}%
\usepackage{amsmath,amssymb,amsfonts}%
\usepackage{amsthm}%
\usepackage{mathrsfs}%
\usepackage[title]{appendix}%
\usepackage{textcomp}%
\usepackage{manyfoot}%
\usepackage{booktabs}%
\usepackage{algorithm}%
\usepackage{algorithmicx}%
\usepackage{algpseudocode}%
\usepackage{listings}%
\usepackage{hhline}
\usepackage{subcaption}
\usepackage{array, tabularx, bm}
\usepackage{adjustbox}
\usepackage{arydshln}
\usepackage{pifont}% http://ctan.org/pkg/pifont
\usepackage{capt-of}
\usepackage{float,wrapfig}
\usepackage{comment}
\usepackage{url}
\usepackage{mathtools, empheq}
\usepackage{enumitem}
\usepackage{makecell}
\usepackage{tcolorbox}
\usepackage{bbding}

\def\ie{\textit{i.e.}}
\def\eg{\textit{e.g.}}

\definecolor{deemph}{gray}{0.6}
\newcommand{\gc}[1]{\textcolor{deemph}{#1}}

\newcommand{\ad}{autonomous driving}
\newcommand{\method}{HiLM-D}
\newcommand{\adapter}{P-Adapter}
\newcommand{\addition}[1]{\color{NavyBlue}
{\footnotesize\textbf{#1}}}
\newcommand{\subtraction}[1]{\color{Peach}
{\footnotesize\textbf{(#1)}}}
\definecolor{mygray}{gray}{.9}
\definecolor{deepgreen}{rgb}{0.0, 0.6, 0.0}
\newcommand{\cmark}{\scalebox{0.9}{\textcolor{deepgreen}{\ding{52}}}}
\newcommand{\xmark}{\textcolor{red}{{\ding{55}}}}

\newcommand{\rev}[1]{{#1}}

\begin{document}

\title[Article Title]{\method: Enhancing MLLMs with Multi-Scale High-Resolution Details for Autonomous Driving}


\author[1]{\fnm{Xinpeng} \sur{Ding}}\email{xdingaf@connect.ust.hk}

\author[2]{\fnm{Jianhua} \sur{Han}}\email{hanjianhua4@huawei.com}

\author[2]{\fnm{Hang} \sur{Xu}}\email{xu.hang@huawei.com}

\author[2]{\fnm{Wei} \sur{Zhang}}\email{wz.zhang@huawei.com}

\author*[1]{\fnm{Xiaomeng} \sur{Li}}\email{eexmli@ust.hk}

\affil[1]{\orgdiv{Department of Electronic and Computer Engineering}, \orgname{The Hong Kong University of Science and Technology}, \orgaddress{\city{Hong Kong SAR}, \country{China}}}

\affil[2]{\orgdiv{Noah’s Ark Lab}, \orgname{Huawei}, \orgaddress{\city{Shanghai}, \country{China}}}

\abstract{
Recent efforts to use natural language for interpretable driving focus mainly on planning, neglecting perception tasks.
In this paper, we address this gap by introducing ROLISP (Risk Object Localization and Intention and Suggestion Prediction), which towards interpretable risk object detection and suggestion for ego car motions.
Accurate ROLISP implementation requires extensive reasoning to identify critical traffic objects and infer their intentions, prompting us to explore the capabilities of multimodal large language models (MLLMs).
However, the limited perception performance of CLIP-ViT vision encoders in existing MLLMs struggles with capturing essential visual perception information,~\eg, high-resolution, multi-scale and visual-related inductive biases, which are important for \ad.
Addressing these challenges, we introduce \method, a resource-efficient framework that enhances visual information processing in MLLMs for ROLISP.
Our method is motivated by the fact that the primary variations in autonomous driving scenarios are the motion trajectories rather than the semantic or appearance information (\eg, the shapes and colors) of objects.
Hence, the visual process of \method~is a two-stream framework: (i) a temporal reasoning stream, receiving low-resolution dynamic video content, to capture temporal semantics, and (ii) a spatial perception stream, receiving a single high-resolution frame, to capture holistic visual perception-related information.
The spatial perception stream can be made very lightweight by a well-designed \adapter, which is lightweight, training-efficient, and easily integrated into existing MLLMs.
Experiments on the DRAMA-ROLISP dataset show \method's significant improvements over current MLLMs,
with a $3.7\%$ in BLEU-4 for captioning and $8.7\%$ in mIoU for detection. 
Further tests on the Shikra-RD dataset confirm our method's generalization capabilities.
}

\keywords{Large language model, Multimodal data, Autonomous driving, Detection}

\maketitle

\section{Introduction}
Over the past decade, there has been remarkable growth in the field of autonomous driving, encompassing both academia and industry~\cite{Singh_Saini_2021}
Generally, the modern autonomous driving system integrates a range of tasks including perception, prediction and planning~\cite {caesar2020nuscenes}.
With the success of deep learning~\cite{He_Zhang_Ren_Sun_2016}, data-driven learning-based methods have become a widespread component of modern autonomous driving systems~\cite{ngiam2021scene,hu2023planning}.
By using sensor data as input,~\eg, RGB or Lidar data, the end-to-end autonomous driving models can directly predict planning or controls for vehicles. 
However, as a black box, these methods generally predict a single score, lacking interpretability and being hard to interact with humans~\cite{deruyttere2019talk2car}.

Recently, some researchers have explored to use a unified natural language for interpretable end-to-end \ad~\cite{deruyttere2019talk2car,xu2023drivegpt4,dewangan2023talk2bev,jin2023adapt,Kim_Misu_Chen_Tawari_Canny_2019}.
However, these approaches mainly focus on planning tasks,~\eg, giving the predictions and reasons for vehicle actions.
To explore the perception tasks, DRAMA~\cite{malla2023drama} is proposed to conduct joint risk localization and description,~\ie, discriminates the most important traffic objects (e.g., cars, traffic cones and so on), explains why the found object is the risk as well as give the bounding box for the found object.
In this paper, we go a further step and extend DRAMA to DRAMA-ROLISP (Risk Object Localization and Intention and Suggestion Prediction), including the additional planning description.
As shown in Table~\ref{tab:comparison}, our task performs more comprehensive tasks compared with current interpretable \ad~tasks.

\begin{table}
\centering
\caption{\textbf{Comparison of our proposed DRAMA-ROLISP with existing language-based driving datasets.} `Percept', `Reason', `Plan' and `Det' indicate the perception, reasoning, planning and detection tasks respectively.}
% \adjustbox{max width=\linewidth}{
  % \scalebox{0.8}{
\begin{tabular}{l|cccc}
\hline
\rowcolor{mygray} Tasks & { \it \textbf{Percept}}& {\it \small Reason} & {\it \small Plan} & {\it \small Det} \\
\hline
\hline
BDD-X~\cite{kim2018textual}  & \xmark & \cmark &  \cmark & \xmark \\
Talk2Car~\cite{deruyttere2019talk2car} & \cmark & \cmark & \xmark & \xmark\\
DRAMA~\cite{malla2023drama} & \cmark & \cmark& \xmark & \cmark \\
Nuscenes-QA~\cite{qian2023nuscenes} &  \cmark & \xmark& \xmark & \xmark \\
DriveGPT4~\cite{xu2023drivegpt4} & \cmark & \cmark & \cmark & \xmark \\
\hline
Ours & \cmark &  \cmark &  \cmark&  \cmark \\
\hline
\end{tabular}
% }
\label{tab:comparison}
\end{table}

To perform accurate ROLISP, it is imperative for the network to have strong reasoning capabilities,~\ie, be able to analyze and discern which traffic object from the current driving scene has the greatest impact on the ego-vehicle, then infer the next intention.
Recently, multimodal large language models (MLLMs) have demonstrated remarkable reasoning abilities in addressing many multimodal tasks~\cite{liu2023visual,zhu2023minigpt, Dai2023instruct}. 
Specifically, they generally align the vision encoder to the large language models (LLMs) by instruct-tuning on image/video-text pairs, thus enabling the LLMs to analyze the context of images or videos following human instructs~\cite{zhang2023video,zhu2023minigpt,li2023videochat}. 
In this paper, we aspire to leverage the powerful ability of MLLMs to address ROLISP task in autonomous driving system.

\begin{figure}[t]
    \centering
\includegraphics[width=\linewidth,height=0.26\textheight]{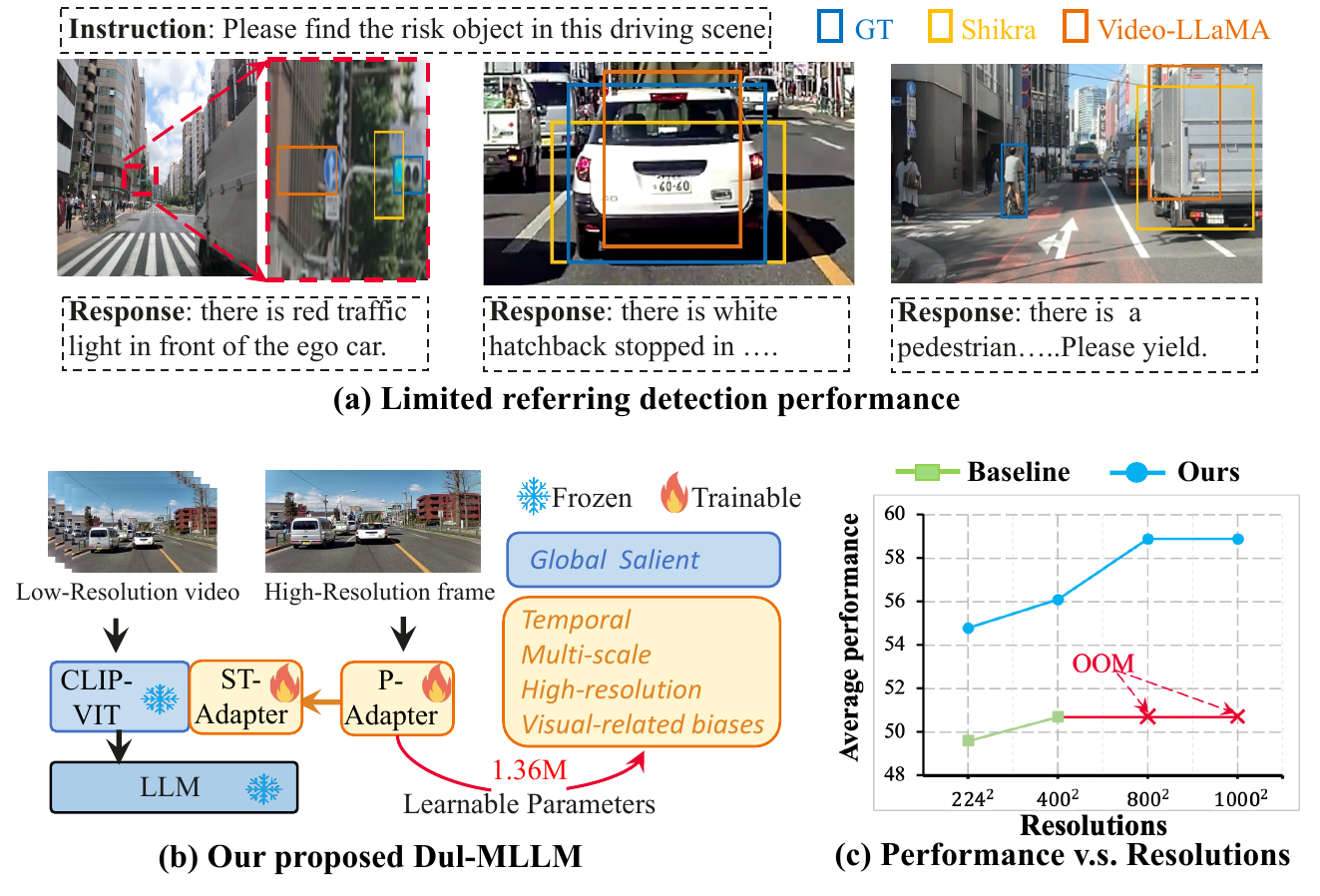}
    \caption{\textbf{(a) Limited referring detection performance.}
    Current MLLMs suffer from failure detection of small objects (left), inaccurate detection of large objects (middle) and over-attention to salient objects (right).
    \textbf{(b) Our proposed \method.} Pre-trained CLIP-ViT~\cite{radford2021learning} only focuses on global salient objects. Using $1.36$M trainable parameters, our method can inject additional information including temporal, multi-scale, high-resolution, and visual-related biases, into existing MLLMs.
    \textbf{(c) Performance v.s. Resolutions.} Our method outperforms the base model (MiniGPT-4~\cite{zhu2023minigpt}) with a clear margin with much less computation and memory cost. See Fig.~\ref{fig:compare} for more experiments.
    }
    \label{fig:motivation}
\vspace{-2.5mm}
\end{figure}
% \vspace{-5mm}

Although great comprehension ability, the limited visual perception performance of vision encoders,~\ie, CLIP-ViT~\cite{radford2021learning}, in existing MLLMs would suffer from several problems when handling \ad,~\eg, missing small objects, generating imprecise bounding box and misidentification risks (refer to Fig.\ref{fig:motivation} for examples).
The reason is that the CLIP-ViT only focuses on the primary visual contents which are just enough for the contrastive training~\cite{li2023fine}.
We argue that in order to process ROLISP, the vision encoder should capture the following information:
\textbf{(i) Temporal cues.} To accurately determine which object poses a risk, it is essential to consider the motion of the current object across consecutive frames.
\textbf{(ii) Multi-scale information.}  In autonomous driving, the sizes of objects vary significantly, ranging from small traffic cones to large trucks, single scale information would miss objects.
\textbf{(iii) Visual-related inductive biases.} Lacking vision-specific inductive biases (~\eg, local connectivity and spatial invariance) would result in slower convergence and lower performance in object detection~\cite{Chen_Duan_Wang_He_Lu_Dai_Qiao_2022,wang2022pvt}.
Solutions that using the advanced vision encoder, such as PVT~\cite{wang2022pvt}, VIVIT~\cite{arnab2021vivit}, DinoV2~\cite{oquab2023dinov2}, would demand significant training resources and still face challenges in capturing all informations,~\eg, temporal, multi-scale, high-resolution information and visual-related inductive biases.

We introduce \method, a resource-efficient approach to capture whole visual information within existing MLLM frameworks for ROLISP.
Compared to training a new vision encoder, \method~utilizes several taloired modules to incorporate temporal, multi-scale information and visual-related inductive biases into CLIP-ViT~\cite{radford2021learning}, as illustrated in Fig.~\ref{fig:motivation}~(b).
\emph{Our main motivation is that in autonomous driving scenarios, the primary variations occur in the motion trajectories rather than the semantic or appearance information of objects. For example, the shapes and colors of vehicles surrounding the ego vehicle remain relatively constant, while only their motion states change.}
Hence, our \method~performs in a two-stream paradigm: a temporal reasoning stream for temporal cues from low-resolution videos, and a spatial perception stream for multi-scale and visual-related inductive biases from high-resolution images.
To this end, the temporal reasoning stream consists of a static visual encoder enhanced with the trainable lightweight ST-Adapters~\cite{pan2022parameter} to enable a pre-trained CLIP-ViT without temporal knowledge to reason about dynamic video content efficiently.
The spatial perception stream has the following parts: a high-resolution spatial encoder to extract fine-grained semantics with visual inductive biases from one high-resolution image, a \adapter~to capture and inject the multi-scale information into the temporal reasoning stream, mining the complete information for the LLM to perform ROLISP. 
Note that our spatial perception stream is very lightweight and training-efficient, which can also act as a plug-and-play module easily integrated into existing MLLMs.

We conduct experiments on the ROLISP benchmark to demonstrate the superiority of \method,~\eg, outperforming the state-of-the-art MLLMs by $3.7\%$ on BLEU-4 for captioning and $8.7\%$ on mIoU for detection with only $1.36$M trainable parameters. Experiments on the generic dataset Shikra-RD~\cite{chen2023shikra} further prove the generalization of our method.

\rev{In summary, the main contributions of our method are as follows:}
\rev{
\begin{itemize}
    \item We introduce ROLISP, a unified task including the risk object localization and Intention and Suggestion Prediction for interpretable autonomous driving.
    \item  We propose a spatial perception stream to capture and inject multi-scale fine-grained details with vision-specific biases into existing MLLMs for autonomous driving, without requiring additional pre-training to align the vision encoder with LLMs.
    \item We introduce a query-aware detector that leverages LLM hidden states as queries to localize objects in high-resolution feature maps, significantly improving object detection accuracy for MLLMs.
    \item We conduct extensive experiments on the ROLISP, trajectory prediction and general tasks to demonstrate the superiority and generalization of our method.
\end{itemize}
}

\definecolor{mygray}{gray}{.9}

%---------------------------------------------
\section{Related Work}
\subsection{Multimodal Large Language Models (MLLMs).}
Natural language processing has witnessed significant strides with the advent of Large Language Models (LLMs)~\eg, GPT series~\cite{OpenAI_2023,radford2019language}, T5~\cite{raffel2020exploring}, LLaMA~\cite{touvron2023llama} and etc. 
Motivated by the potential of LLMs, numerous multimodal LLMs (MLLMs),~\eg, LLaVA~\cite{liu2023visual}, MiniGPT-4~\cite{zhu2023minigpt}, Video-LLaMA~\cite{zhang2023video} and InstructBLIP~\cite{Dai2023instruct}, have been proposed to expand the LLMs to the multimodal field,~\ie, perceiving image/video input, and conversating with users in multiple rounds.
Pre-trained on massive image/video-text pairs, the above models can only handle image-level tasks, such as image captioning and question answering.
Hence, several works,~\eg ContextDET~\cite{zang2023contextual}, KOSMOS-2\cite{peng2023kosmos} and Shikra~\cite{chen2023shikra}, have been proposed to enable the grounding ability of MLLMs to produce bounding boxes.
%
% Specifically, they generally construct the instruction data including bounding box annotations by human or automatic tools~\ie, GPT-4~\cite{OpenAI_2023}, and pre-train the MLLMs  
%
However, all of the current MLLMs train the model in low-resolution image-text pairs, presenting limited perception results in high-resolution autonomous driving scenarios.
%
% To discern fine-grained visual details from high-resolution inputs, an intuitive approach is to split images into patches and project them using linear layers, treating these as a sequence for input into Large Vision-Language Models (LVLMs)~\cite{fuyu-8b,li2023otterhd}. While this eliminates the need for an image encoder, it often results in insufficient visual representation, leading to increased training costs and suboptimal performance.
To capture fine-grained visual details from high-resolution inputs, images are split into patches and projected with linear layers for input into Large Vision-Language Models (LVLMs)~\cite{fuyu-8b,li2023otterhd}, avoiding an image encoder but often resulting in poor visual representation and higher training costs.
Up-Resize methods like Qwen-VL~\cite{Qwen-VL} adjust Vision Transformer (ViT) embeddings from $224 \times 224$ to $448 \times 448$, adding a training phase to fine-tune the ViT. However, this can degrade visual representation by altering the original position encoding~\cite{radford2021learning}.
Slicing-based methods~\cite{li2023monkey,xu2024llava} divide high-resolution images into patches matching a pre-trained vision encoder's input size, maintaining efficiency while achieving competitive performance. However, they demage the original context and spatial continuity across patches, as well as losing visual-related inductive biases~\cite{wang2022pvt}.

% Dual-branch approaches introduce a high-resolution branch with lightweight convolutional networks to manage high-resolution inputs but require additional training data and parameters~\cite{hong2023cogagent,ding2023hilm,luo2024feast,li2024mini}.
%
% Slicing-based methods offer a compromise by using slicing windows to divide the high-resolution image into patches that match the input size of a pre-trained vision encoder, maintaining efficiency in parameter use and training data while still achieving competitive performance~\cite{li2023monkey,xu2024llava}. However, they suffer from "Context Fragmentation", where the continuity of contextual information across patches is damaged, impacting tasks that require cross-patch context and spatial relationships.
%
% In this paper, we propose \method, a novel technique designed to seamlessly integrate global-local high-resolution details into LVLMs without disrupting the original context or spatial geometry, effectively addressing the issue of Context Fragmentation.
%
Differently, our \method~maintain the whole high-resolution details while capturing multi-scale and visual-related inductive biases informations.

\begin{figure*}[t]
    \centering
\includegraphics[width=1\textwidth,height=0.35\textheight]{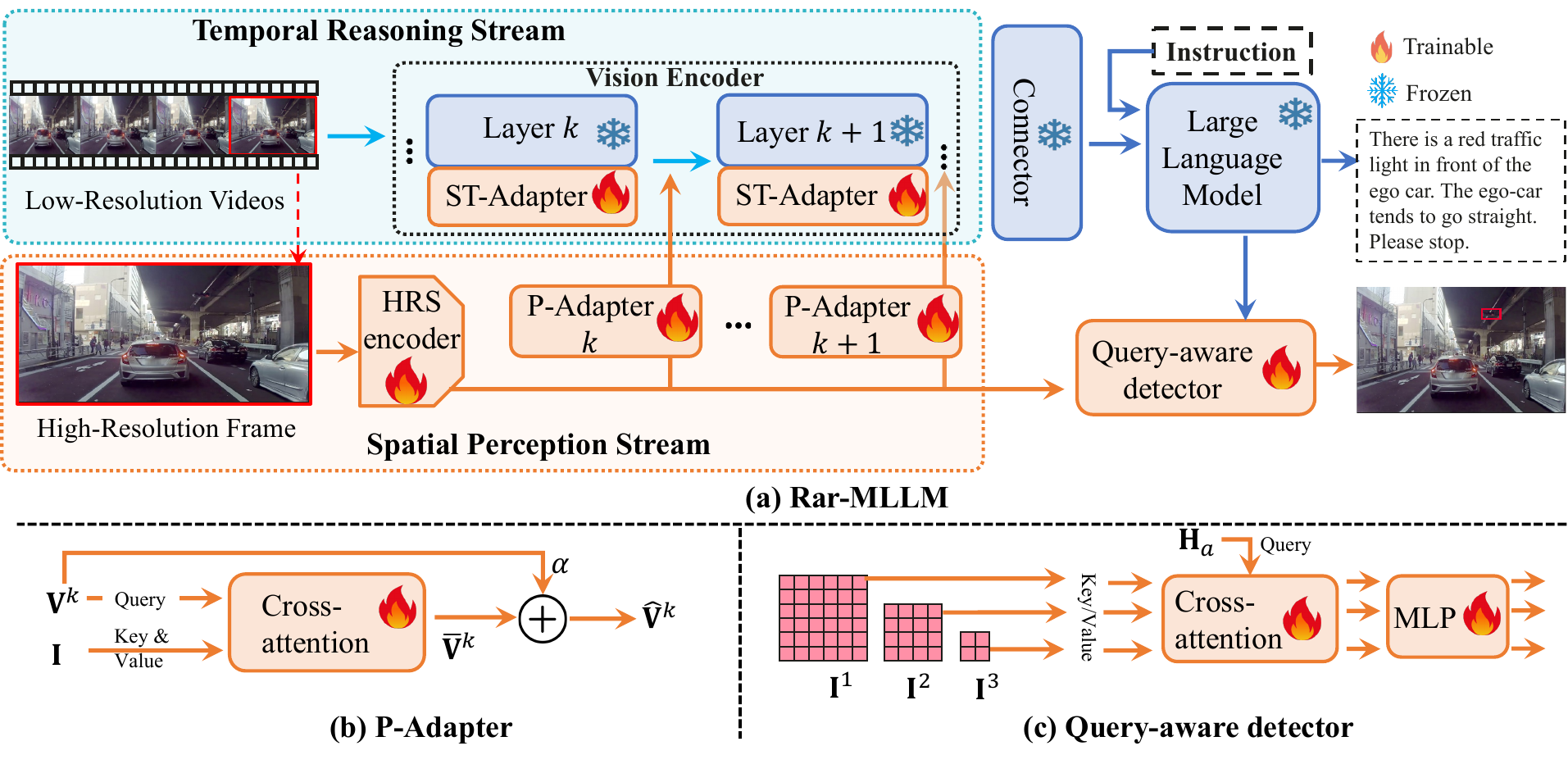}
\vspace{-2.0em}
    \caption{\textbf{(a) Overall pipeline of \method.} The visual process consists of two streams,~\ie, temporal reasoning stream and spatial perception stream, to capture temporal, multi-scale, fine-grained visual information with visual-related inductive biases,~\ie, $\{ \mathbf{I}^i \}_{i=1}^3$. \textbf{(b) \adapter.} Through the cross-attention mechanism, \adapter~incoporates the enhanced multi-scale information,~\ie, $\{ \mathbf{I}^i \}_{i=1}^3$, into the pre-trained vision encoder CLIP-ViT.
  \textbf{(c) Query-Aware Detector.} For imprecise bounding boxes of risk objects, the query-aware detector regard the hidden-states of the LLM as the prior knowledge to refer the bounding box in $\{ \mathbf{I}^i \}_{i=1}^3$.  
    }
    \label{fig:architecture}
\vspace{-2mm}
\end{figure*}

\vspace{-2mm}
\subsection{Risk Object Identification.}
Risk object identification methodologies can be grouped into explicit and implicit learning.
Explicit methods~\cite{zeng2017agent,gao2019goal} use binary classification for agent importance estimation, while approaches~\cite{alletto2016dr,tawari2018learning} mimic human gaze for risk proxy via pixel-level attention maps.
Implicit methods~\cite{malla2020titan,kim2017interpretable,wang2019deep,li2020make} focus on related tasks like trajectory prediction, with intermediate activations indicating perceived risk. 
% Particularly, [26] uses self-attention for future trajectory prediction and agent importance, while [24] adopts a causal intervention approach. Methods in [19, 35] predict steering control and pixel-level attention for agent importance.
However, these methods lack reasoning about model decisions or natural language descriptions,
limiting interpretability in autonomous driving and driver-assistance systems.
Recently, ADAPT~\cite{jin2023adapt} leverages the transformer to generate action narration and reasoning for the intention of self-driving vehicles.
DRAMA~\cite{malla2023drama} proposes a new direction of risk object identification that gives the risk object and explanation in natural languages.
% limiting interpretability in autonomous driving and driver-assistance systems.
%
In this paper, we go one step further than Drama and ADAPT,~\ie, risk object localization and intention prediction (ROLISP) that aims to identify, explain and localize the risk object for the ego-vehicle meanwhile predicting its intention.

\vspace{-2mm}
\subsection{Multi-tasks in Autonomous Driving.}
Traditional autonomous driving algorithms individually process different tasks,~\eg, detection~\cite{chen2017multi}, tracking~\cite{petrovskaya2008model}, reasoning~\cite{kim2018textual}, and prediction~\cite{ngiam2021scene}.
To extract the richer inter-task information, researchers explore integrating the multi-tasks in the end-to-end training frameworks.
For instance, combined training for detection and tracking has been demonstrated in works like D \& T~\cite{petrovskaya2008model}. FaF~\cite{luo2018fast} takes this further by unifying a detector with a trajectory predictor, yielding notable results. 
IntentNet~\cite{casas2018intentnet} expanded this paradigm by also integrating intention prediction for actors. 
UniAD~\cite{hu2023planning} stands out by amalgamating full-stack driving tasks within a singular framework, albeit still relying on distinct sub-networks for each task.
A novel direction in this domain is the use of natural language as a unified output across tasks.
For instance, ADAPT~\cite{jin2023adapt} predicts intention and gives explanations using a single caption, while DRAMA~\cite{malla2023drama} aims for the risk object detection and explanation.
In this paper, we go one step further than Drama and ADAPT,~\ie, ROLISP that aims to identify, explain and localize the risk object for the ego-vehicle meanwhile predicting its intention and giving suggestions.

\section{Method}~\label{Sec:method}
%
% In this section, we present our model that feeds videos with text prompts to output risk localization for autonomous driving,~\ie, identifying the most important object that may influence the ego-vehicle; and captioning why the object is important.
% This section presents our \method~ that receives videos to address ROLISP.
%

\vspace{-5mm}
\noindent This section details our \method~approach designed to tackle ROLISP using video inputs as shown in Fig.~\ref{fig:architecture}.
The vision encoder of \method~consists of two streams, a temporal reasoning stream for temporal cues (Section~\ref{sec:temporal}) and a spatial perception stream for fine-grained multi-scale information as well as visual inductive biases (Section~\ref{Sec:spatialbranch}).
The features from the two streams are performed in a cooperative way to achieve the enhanced visual feature, which is then fed to LLMs to perform risk object identification, reasoning and planning.
Additionally, we propose the query-aware detector for the referred bounding box of the found risk object, illustrated in Section~\ref{sec:detector}.
%
% As depicted in Fig.~\ref{fig:architecture}, our method is end-to-end and consists of two primary components:
% (a) a multimodal large language model (MLLM) branch to process video semantics to generate captions about risk objects, reasons, and ego-car actions.
% %
% % capture high-level temporal semantics of videos to produce captions for the risk object with reasons and the ego-car intention and motion suggestions;
% %
% (b) a high-resolution perception branch (HR-PB) to extract HR feature maps with vision-specific cues from HR images to enhance the perception ability of the MLLM.
% %
% As shown in Fig.~\ref{fig:architecture}, our method is end-to-end and consists of four main components:
% %
% (a) a visual encoder equipped with a temporal adapter to capture high-level features of videos;
% %
% (b) a decoder to produce the captions for the risk object and the corresponding bounding box;
% %
% (c) an efficient spatial adapter that incorporates high-resolution semantics into the high-level video features, enhancing the localization ability.

%
\subsection{Temporal Reasoning Stream}~\label{sec:temporal}

\vspace{-5mm}
\noindent As shown in Fig~\ref{fig:architecture}~(a), the temporal reasoning stream is built on the well pre-trained vision encoder,~\ie, CLIP-ViT~\cite{radford2021learning}, equipped with new learnable spatial-temporal adapters following \cite{pan2022parameter} for video reasoning. 
Formally, given a video with $L$ frames, the CLIP-ViT maps each frame to its $k$-th layer feature, resulting in $\mathbf{V}^k = \{ \mathbf{v}_i^k \}_{i=1}^L$, where $\mathbf{v}_i^k$ is the feature of $i$-th frame, $\mathbf{v}_i^k \in \mathbb{R}^{N_v \times D_v}$, $N_v$ is the patch number and $D_v$ is the dimension.
We then reshape its dimension to $\mathbf{V}^{k\prime} \in \mathbb{R}^{L \times H_v \times W_v \times D_v}$, where $H_v * W_v = N_v$.
Then, to obtain the spatial-temporal information, a standard depth-wise 3D convolution layer~\cite{feichtenhofer2020x3d} is used, which can be formulated as:
\begin{equation}
\begin{split}
    \text{ST-Adapter}(\mathbf{V}^{k\prime})=\mathbf{V}^{k\prime}+ \\f\left(\operatorname{DWConv3D}\left(\mathbf{V}^{k\prime} \mathbf{W}_{\text {down }}\right)\right) \mathbf{W}_\text{up},
\end{split}
\end{equation}
\rev{where} $\operatorname{DWConv3D}$ denotes the depth-wise 3D-convolution, $f$, $\mathbf{W}_{\text {down }}$ and $\mathbf{W}_\text{up}$ are the activation function, down-sampling and up-sampling weight.
Then, we reshape $\text{ST-Adapter}(\mathbf{V}^{k\prime})$ back to $\mathbf{V}^{k}$.

\subsection{Spatial Perception Stream}~\label{Sec:spatialbranch}

\vspace{-3mm}
\noindent \rev{Achieving effective perception requires high-resolution inputs, visual-related inductive biases, and multi-scale, fine-grained information. However, integrating these elements into existing Multimodal Large Language Models (MLLMs) presents a significant challenge.
The primary issue arises from the fact that current MLLMs typically employ CLIP-ViT~\cite{radford2021learning} as the vision backbone, which has been pre-aligned with language embeddings for LLMs. CLIP-ViT is designed with a fixed input resolution, such as $224 \times 224$ or $336 \times 336$, and lacks visual-specific inductive biases that are crucial for tasks like object detection.
Directly training a new high-resolution Vision Transformer (ViT) with visual-related inductive biases, such as Swin Transformers~\cite{liu2021swin}, for video-based tasks would require vast amounts of data, substantial memory, and significant computational resources, resulting in prohibitive costs in both training time and infrastructure.}

In autonomous driving scenarios, the primary variations occur in the motion trajectories rather than the semantic or apperance information of objects. For example, the shapes and colors of vehicles surrounding the ego vehicle remain relatively constant, while only their motion states change, and low-resolution input is sufficient to capture these motion trajectories. Motivated by this, we sample a single frame from the original video in high-resolution to extract the detailed perceptual information (In this paper, we use sampled the last frame from the video; see Table~\ref{tab:frame} for analysis).

To this end, we propose a spatial perception stream, which captures and integrates multi-scale vision-specific information from the high-resolution image into the MLLM.
%
% The high-resolution perception branch (HR-PB) is tailored to integrate vision-specific information from the high-resolution images and features associated with potential high-risk objects into the MLLM.
%
As shown in Fig.~\ref{fig:architecture}~(a), the proposed stream consists of two main parts: a high-resolution spatial (HRS) encoder to obtain the multi-scale features with visual-specific biases from the high-resolution frame and an \rev{\adapter~to} incorporate the extracted features into the temporal reasoning stream.

\noindent\textbf{HRS encoder.} To capture multi-scale vision-specific information for object detection, the HRS encoder is modified from the classic convolution network (CNN) ResNet~\cite{He_Zhang_Ren_Sun_2016}.
Compared with the plain ViT in current MLLMs, CNN incurs many advantages: reducing the memory and computation resources, and bringing vision-specific prior for the detection tasks (\eg, local connectivity and spatial invariance).
We prove this in Table~\ref{tab:HRES}.
For multi-scale features, we use a stack of stride-2 $3 \times 3$ convolution layers to double the number of channels and reduce the size of feature maps, resulting in hierarchical features $\{ \mathbf{I}^i \}_{i=1}^3$ with resolutions of $1/8$, $1/16$, and $1/32$.
%
% We denote the extracted hierarchical features as $\{ \mathbf{I}^i \}_{i=1}^3$, where $\mathbf{I}^i 
%  \in \mathbb{R}^{(H*W)/2^{i+2} \times D_i}$, $D_i$, $H$ and $W$ are the dimension, width and height respectively.
%
Then, we flatten and concatenate these feature maps into feature tokens $\mathbf{I} \in \mathbb{R}^{N_{sp}\times D_i}$, where $N_{sp}$ and $D_i$ are the token number and the dimension respectively.
% %

\noindent\textbf{\adapter.}
The \rev{\adapter~aims} to incorporate the multi-scale high-resolution spatial features into the features from CLIP-ViT~\cite{radford2021learning}, thus allowing the LLM to receive the complete visual information to compare and decide which object needs the most attention.
As shown in Fig.~\ref{fig:architecture}~(b), for the $k$-th block of the vision encoder, X-Adapter take the temporal low-resolution visual embedding $\mathbf{V}^k$ (See Section~\ref{sec:temporal}) as query, and the extracted multi-scale spatial visual features,~\ie, $\mathbf{I}$ (See Section~\ref{Sec:spatialbranch}) as key and value.
Then, the \adapter~process can be formulated as:
\begin{equation}
    {\mathbf{V}}^{k+1}= \mathbf{V}^k + \alpha \text{Cross-Attn}(\text{norm}(\mathbf{V}^k),\text{norm}(\mathbf{I})),
    \label{e:xadapter}
\end{equation}
where $\text{norm}(\cdot)$ and Cross-Attn are LayerNorm and cross-attention layer respectively.
$\alpha$ is a learnable gating factor to adaptively control the importance of enhanced features and $\mathbf{V}^k$.
We initialize $\alpha$ as zero to ensure the feature distribution of pre-trained CLIP-ViT will not be modified drastically, for better knowledge maintaining and stable optimization~\cite{zhang2023llamaadapter}.

\rev{Note that although our method adopts a two-stream framework, the motivation and technical details differ from the previous video two-stream framework (such as Slow-Fast~\cite{feichtenhofer2019slowfast}).}
\rev{Existing two-stream frameworks like Slow-Fast focus on {temporal dynamics} for general video semantic classification, typically using low-resolution frames (224$\times$224) with varying temporal resolutions (e.g., Slow-Fast~\cite{feichtenhofer2019slowfast} uses different frame rates and others~\cite{simonyan2014two,carreira2017quo} use the additional optical flow stream).}
\rev{In contrast, our method targets {both temporal dynamics and fine-grained spatial perception}, crucial for tasks like autonomous driving. Our framework achieves this by introducing additional lightweight modules to capture holistic perception semantics and efficiently inject them into MLLM without the need for massive pre-training or architecture modification,~\eg using only 1.36M trainable parameters to achieve $8.7\%$ improvements on detection tasks.}

\subsection{Large Language Model}
% \noindent \textbf{Large Language Model (LLM).}
%
Given the visual tokens and text instruction tokens, the pre-trained LLM (~\eg, Vicuna~\cite{chiang2023vicuna}) is leveraged for descrition generation, including the risk object with explanations, intentions and suggestions for the ego-car.
The input to the LLM is the concatenated multimodal tokens $[{\mathbf{Z}_v}, \mathbf{Z}_t] \in \mathbb{R}^{(N_v+N_t) \times D_t}$, where $\mathbf{Z}_v = \text{project}(\mathbf{V}^K)$ and $\text{project}(\cdot)$ is the projection layer to transfer the visual embedding into tokens that LLM can understand. $\mathbf{Z}_t \in \mathbb{R}^{N_t \times D_t}$ are the text embeddings, tokenized from text prompts,~\eg, \texttt{`Which object is at the highest risk? Then predict the motions and suggestions for the ego-car'}.
Then, the pre-trained LLM recevies the mulimodal tokens $[{\mathbf{Z}_v}, \mathbf{Z}_t]$ to generate language in an autoregressive way as follows:
\begin{equation}
p\left(\mathbf{Z}_a \mid \mathbf{Z}_v, \mathbf{Z}_t \right)=\prod_{i=1}^L p_{{\theta}}\left(\mathbf{z}_i \mid \mathbf{Z}_v, \mathbf{Z}_{t,<i}, \mathbf{Z}_{a,<i} \right),
\end{equation}
where $\theta$ is the trainable parameters, $\mathbf{Z}_a \in \mathbb{R}^{N_a \times D_t}$ is the generated answer, $\mathbf{Z}_{t,<i}$ and $\mathbf{Z}_{a,<i}$ are the prompts and answer tokens before the current prediction token $\mathbf{z}_i$.
The ST-Adapters and the linear layer are supervised by the caption loss $\mathcal{L}_{cap} = CE(\mathbf{Z}_a, \hat{\mathbf{Z}}_a)$, where $\hat{\mathbf{Z}}_a$ is the ground-truth answer, $CE$ is the cross-entropy loss.

\subsection{Query-aware Detector}~\label{sec:detector}

\vspace{-3mm}
\noindent \rev{For} obtaining the precise bounding box of the identified risk object, we devise a query-aware detector to regard the hidden states as the prior knowledge to find the bounding box in the multi-scale visual features~\ie, $\{ \mathbf{I}^i \}_{i=1}^3$.
% queries the hidden states from 
%
Normally, the last token of the hidden-states fully perceive the whole multimodal context and contains comprehensive instruction-aware semantics~\cite{li2023fine}.
Therefore, we use the hidden states of the last token, $\mathbf{H}$, as the query, referring the visual cues from $\{ \mathbf{I}^i \}_{i=1}^3$ for bounding boxes.  
This process can be presented as follows:
\begin{equation}
    \overline{\mathbf{H}}_i = \text{Cross-Attn}(\mathbf{H},\mathbf{I}_i).
    \label{e:query}
\end{equation}
Finally, for each scale, $\overline{\mathbf{H}}_i$ is fed to two MLP layers to generate the bounding box $B$ and the activation score $c$, 
\begin{equation}
    B = \text{MLP}(\overline{\mathbf{H}}_i),
    c = \text{Softmax}(\text{MLP}(\overline{\mathbf{H}}_i)).
\end{equation}
During training, we define a multi-task loss to supervise the bounding box prediction as $\mathcal{L}_\text{det} = \mathcal{L}_\text{act} + \mathcal{L}_\text{box}$, where $\mathcal{L}_\text{act}$ is the cross-entropy loss and $\mathcal{L}_\text{box}$ is the L1 loss.
The overall loss is defined as follows:
\begin{equation}
    \mathcal{L} = \mathcal{L}_\text{cap} + \lambda_\text{det}\mathcal{L}_\text{det},
    \label{E:fullloss}
\end{equation}
where $\lambda_\text{det}$ is the hyper-parameters.

%----------------------------------------------------

%--------------------------------
\begin{figure}[t]
    \centering
    \includegraphics[width=0.9\linewidth,height=0.4\textheight]{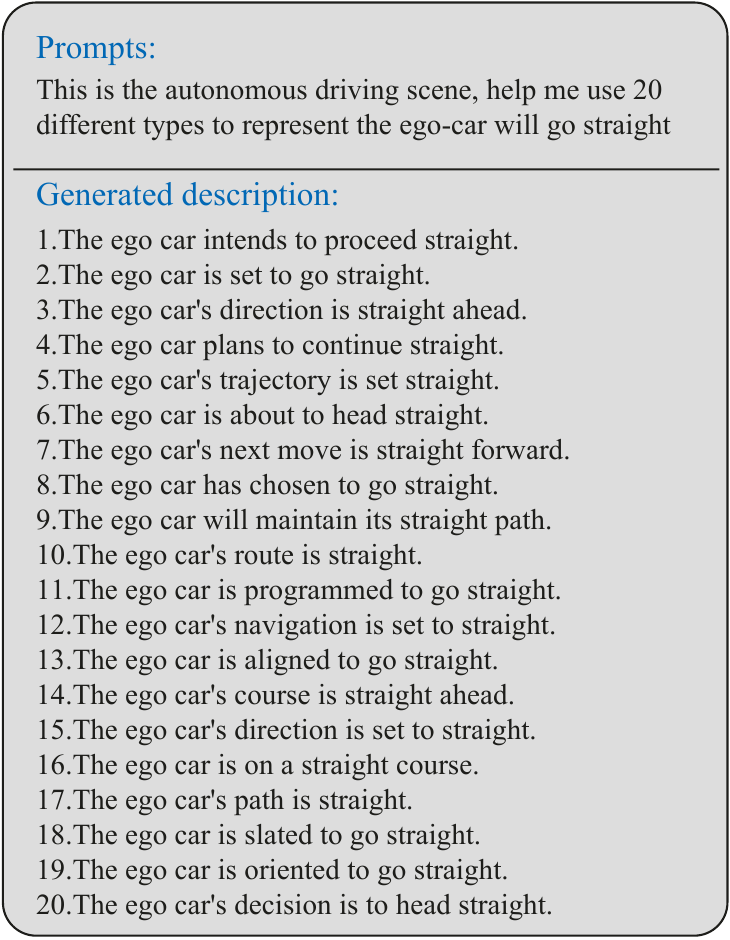}
    \caption{\textbf{Examples of generated descriptions by GPT-4~\cite{OpenAI_2023}.}  
    }
    \label{fig:generate_example}
\end{figure}
%------------------------------------
%
%--------------------------------
\begin{figure}
    \centering
    \includegraphics[width=\linewidth,height=0.18\textheight]{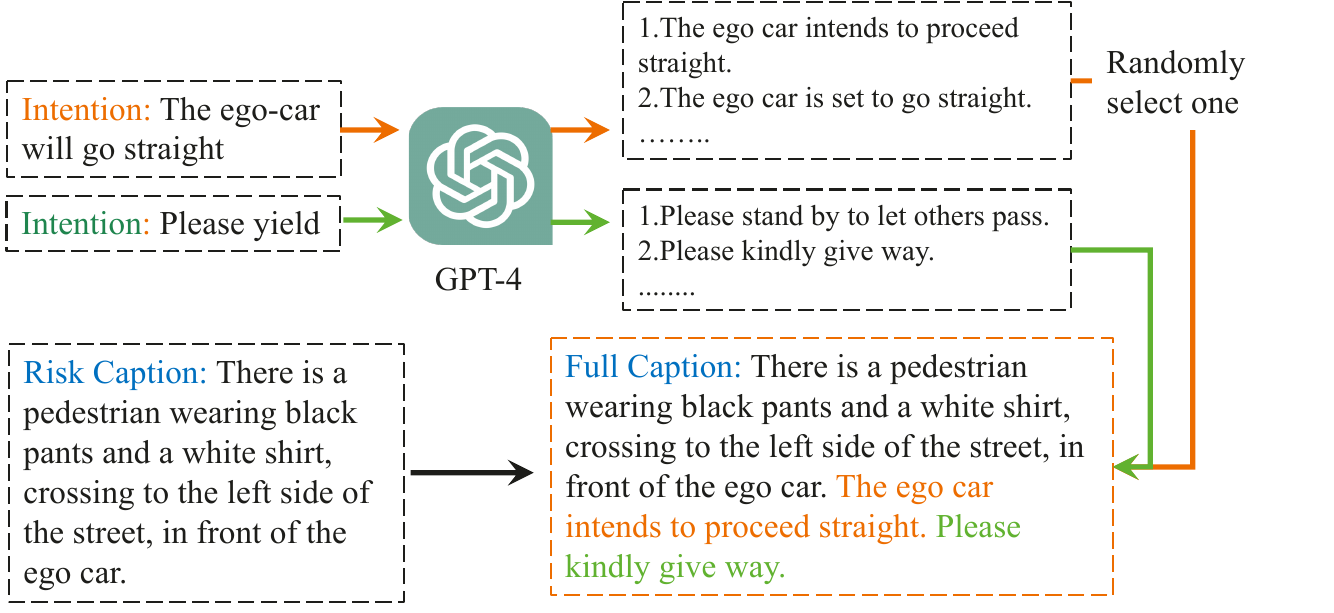}
    \caption{\textbf{Illustration of the caption construction for ROLISP.}  
    }
    \label{fig:construct}
\end{figure}
%------------------------------------

\definecolor{mygray}{gray}{.9}

\section{Experiments}

%------------------------------------

\subsection{Experimental Settings}~\label{sec:experimentsetting}

\noindent \rev{\textbf{Architecture Details.} In this paper, we adopt the well-known MLLM, MiniGPT-4\cite{zhu2023minigpt}, as our backbone. Notably, our proposed \adapter~is a plug-and-play module that can be seamlessly integrated into existing MLLMs (see Table~\ref{tab:applied}). As described in Section~\ref{Sec:spatialbranch}, the HRS encoder is a modified version of the classic CNN-based ResNet~\cite{He_Zhang_Ren_Sun_2016}. Specifically, it consists of an initial 7×7 convolutional block followed by four residual stages (64→128→256→512) to extract spatial features. The query-aware detector comprises a cross-attention head (a multi-head attention block) and three MLP layers, each followed by a ReLU activation, to map features to bounding boxes.}

\noindent \textbf{Implementation details.}
Our proposed method is implemented in PyTorch~\cite{paszke2019pytorch} trained using a single machine with $8$ NVIDIA V100 GPUs.
The input video frames are resized and cropped to the spatial size of $224 \times 224$.
We uniformly sample $L=5$ frames from the entire video, and ensure that the last frame is sampled for producing bounding boxes.
We set  $\lambda_\text{det}$ in Eq.~\ref{E:fullloss} to $2$.
We use AdamW~\cite{loshchilov2017decoupled} as the optimizer and cosine annealing scheduler~\cite{loshchilov2016sgdr} as the learning rate scheduler with an initial learning rate of $1e$-$4$ for the MLLM branch and $4e$-$4$ for the spatial perception stream, and global batch size of $64$.
%

% \subsection{Datasets and Evaluation Metrics}
%

\noindent \textbf{Datasets.}
DRAMA~\cite{malla2023drama} is a benchmark evaluating visual reasoning in driving scenarios with $17,785$ two-second interactive scenarios.
The videos in DRAMA~\cite{malla2023drama} dataset are recorded in $30$hz frame rate, $1928 \times 1280$ resolution or $60$hz frame rate, $2704 \times 1520$ resolution.
We split the total $17,785$ videos into $70\%$ train, $15\%$ validation and $15\%$ test.
The original captions for DRAMA only contain the risk object and its explanation.
To add the intention and suggestions for the ego-car, we use the answer based on the video question answer (VQA) annotations from the dataset to complete the captions for ROLISP (Risk Object Localization and
Intention and Suggestion Prediction).
%
% The VQA annotations from DRAMA are shown in Fig.~\ref{fig:dataset}~(b).
%
We obtain the ground truth of intentions and suggestions for the ego-car by the question: `Suggestions (to the ego car)' and `Intention of Ego-vehicle'.
However, the intention and suggestion answers are shot and single,~\ie, `straight', `yield'.
Hence, we resort to GPT~\cite{OpenAI_2023} to generate more diverse descriptions.
For example, we use the prompt as follows:
\begin{tcolorbox}
{This is the autonomous driving scene, help me use 20 different types to represent the ego-car will go straight.}
\end{tcolorbox}
Fig.~\ref{fig:generate_example} illustrates the newly generated descriptions. 
Similarly, we can also obtain the descriptions for `yield'.
We randomly select one of the 20 generated descriptions and add it to the original risk object's caption in the order of intention and suggestion, obtaining the full caption for ROLISP.
The whole process is illustrated in Fig.~\ref{fig:construct}.
% \begin{mdframed}[style=MyFrameStyle]

\begin{table*}
\centering
\caption{ \textbf{Comparison with the state-of-the-art.}
For all metrics, the higher the scores, the better the results. 
`Ours w/o ST' means our model without ST-Adapters. 
`B4', `M', `C' and `S' refer to BLEU-4, METEOR, CIDER and SPICE.
\rev{`LLM' means using the LLM as a judge for semantic correctness (see details in Section~\ref{sec:experimentsetting}).}
`AVG' is the average value of B4 and mIoU.
All results are obtained by fine-tuning the models on DRAMA-ROLISP in the same setting.
The entries noted in grey are the results of our method.
$*$ indicates that the model is pre-trained on large object detection datasets. 
%
% \vspace{-1.0em}
}
% \begin{footnotesize}
  \begin{adjustbox}{max width=\linewidth}
  \begin{tabular}{c|l|l|ccccc|cccc|c}
    \toprule
    % \Xhline{1.0pt}
  {\multirow{2}{*}{Input}} & {\multirow{2}{*}{Method}} & {\multirow{2}{*}{Resolution}} & \multicolumn{5}{c|}{\footnotesize \it Captioning}  & \multicolumn{4}{c|}{\footnotesize \it Detection} &  {\multirow{2}{*}{AVG}} \\
%      % \cmidrule{3-6} \cmidrule{8-8} 
   & &  & B4 & M  & C &  S & \rev{LLM}  & mIoU  & IoU$_{S}$  & IoU$_{M}$  &IoU$_{L}$  &  \\
%     \\
     % \cmidrule{1-11}
     \hline
     % \midrule
% %     % \cmidrule{1-2} \cmidrule{3-6} \cmidrule{6-11}
 \multirow{5}{*}{Image}   & BLIP-2~\cite{li2023blip} &  $224 \times 224$  & 46.1 &  34.3 &194.7 & 50.7 & \rev{3.02} & 46.3 & 8.1 &  60.2  & 73.7 &  46.2\\
   % & MiniGPT-4 & \\
   % & LLaVA~\cite{liu2023visual} & 224 &47.5 &  35.2 & 198.6 & 48.3 & 47.2 & 8.0 &  62.1  & 74.2 & 47.4\\
  & InstructBLIP~\cite{Dai2023instruct} & $224 \times 224$ & 49.9 &  37.9 & 205.0 & 50.9 & \rev{3.14} & 47.8 & 9.1 &  62.2  & 74.5 & 48.9\\
 & MiniGPT-4~\cite{zhu2023minigpt} &  $224 \times 224$  & 50.5 &  36.1 & 208.8 & 49.2 &  \rev{3.27} & 46.3 & 9.0 &  63.1  & 72.2 &48.4\\
  & Shikra$^*$~\cite{chen2023shikra}&  $224 \times 224$  &49.8 &  37.7 & 204.7 & 50.7 & \rev{3.28} & 50.3 & 10.4 &  59.5  & 73.8 & 50.1 \\
  & LLaVA-1.5~\cite{liu2023visual} &  $336 \times 336$  & 50.3 &  38.5 & 213.6 & 52.3 & \rev{3.31} & 47.2 & 8.0 &  62.1  & 74.2 & 48.8\\
  & Monkey~\cite{li2023monkey} & $1344 \times 896$ & 53.2 &  37.4 & 210.7 & 50.5 & \rev{3.39} & 51.8 & 15.5 &  60.5  & 76.8 & 52.5\\
  & \rev{Qwen-VL~\cite{Qwen-VL}} & \rev{$448 \times 448$} & \rev{54.5} &  \rev{38.2} & \rev{218.6} & \rev{51.3} & \rev{3.41} & \rev{50.8} & \rev{12.8} &  \rev{58.7}  & \rev{77.9} & \rev{52.7}\\
  & \rev{LLaVA-Next~\cite{liu2024llavanext}} & \rev{$672 \times 672$} & \rev{55.2} &  \rev{38.5} & \rev{220.6} & \rev{52.2} & \rev{3.38} & \rev{49.5} & \rev{13.8} &  \rev{57.2}  & \rev{74.5} & \rev{52.3}\\
& \cellcolor[gray]{0.9}Ours w/o ST  & \cellcolor[gray]{0.9} $1000 \times 1000$  & \cellcolor[gray]{0.9}\textbf{56.3} & \cellcolor[gray]{0.9} \textbf{40.0} & \cellcolor[gray]{0.9} \textbf{238.1}& \cellcolor[gray]{0.9} \textbf{54.5} & \cellcolor[gray]{0.9} \rev{\textbf{3.50}} &\cellcolor[gray]{0.9} \textbf{60.5}& \cellcolor[gray]{0.9} \textbf{33.5} & \cellcolor[gray]{0.9} \textbf{64.1}& \cellcolor[gray]{0.9}\textbf{81.8} & \cellcolor[gray]{0.9}\textbf{58.4}\\
  % \cmidrule{1-11}
   % \midrule
  \hline
\multirow{3}{*}{Video}%
& eP-ALM~\cite{shukor2023ep} & $224 \times 224$ & 51.4 & 38.0 & 225.1 & 52.8 & \rev{3.36} & 43.2 & 7.2 & 56.8 & 68.8 & 47.3 \\
  & Video-LLaMA~\cite{zhang2023video} & $224 \times 224$  & 53.9 & 37.8 & 229.5 & 52.6 & \rev{3.42} & 42.8 & 6.9 & 55.3 & 67.9 & 48.4 \\
  & \rev{Video-LLaMA 2~\cite{damonlpsg2024videollama2}} & \rev{$336 \times 336$}  & \rev{54.6} & \rev{38.9} & \rev{234.8} & \rev{53.7} & \rev{3.44} & \rev{45.2} & \rev{7.6} & \rev{57.8} & \rev{70.3} & \rev{49.9} \\
 & \cellcolor[gray]{0.9} Ours & \cellcolor[gray]{0.9} $1000 \times 1000$ & \cellcolor[gray]{0.9}\textbf{58.3} & \cellcolor[gray]{0.9}\textbf{41.6} & \cellcolor[gray]{0.9}\textbf{278.5}& \cellcolor[gray]{0.9}\textbf{57.5} &\cellcolor[gray]{0.9} \rev{\textbf{3.62}} & \cellcolor[gray]{0.9}\textbf{59.6}& \cellcolor[gray]{0.9}\textbf{32.4} & \cellcolor[gray]{0.9}\textbf{62.8}&\cellcolor[gray]{0.9}\textbf{82.5} &\cellcolor[gray]{0.9}\textbf{58.9} \\
\hline
\gc{\multirow{2}{*}{Specialist SOTAs}} & \gc{ADAPT~\cite{jin2023adapt}} &   \gc{$224 \times 224$} & \gc{57.9} & \gc{38.9} & \gc{240.6} & \gc{53.5} & \gc{\rev{3.48}} & \gc{-} &  \gc{-} & \gc{-} & \gc{-} & \gc{-} \\
 & \gc{Grounding-Dino~\cite{liu2023grounding}} &  \gc{$1000 \times 1000$}  & \gc{-} &  \gc{-} & \gc{-}& \gc{-} & \gc{-} & \gc{58.6} & \gc{2.6} & \gc{70.5} & \gc{93.2}  & \gc{-}   \\
% \bottomrule
\hline
    \end{tabular}
    \end{adjustbox}
    \label{tab:sota}
    \vspace{-3mm}
% \end{footnotesize}
\end{table*}
%----------------------------------------------------

\begin{figure}
    \centering
    \includegraphics[width=\linewidth,height=0.45\textheight]{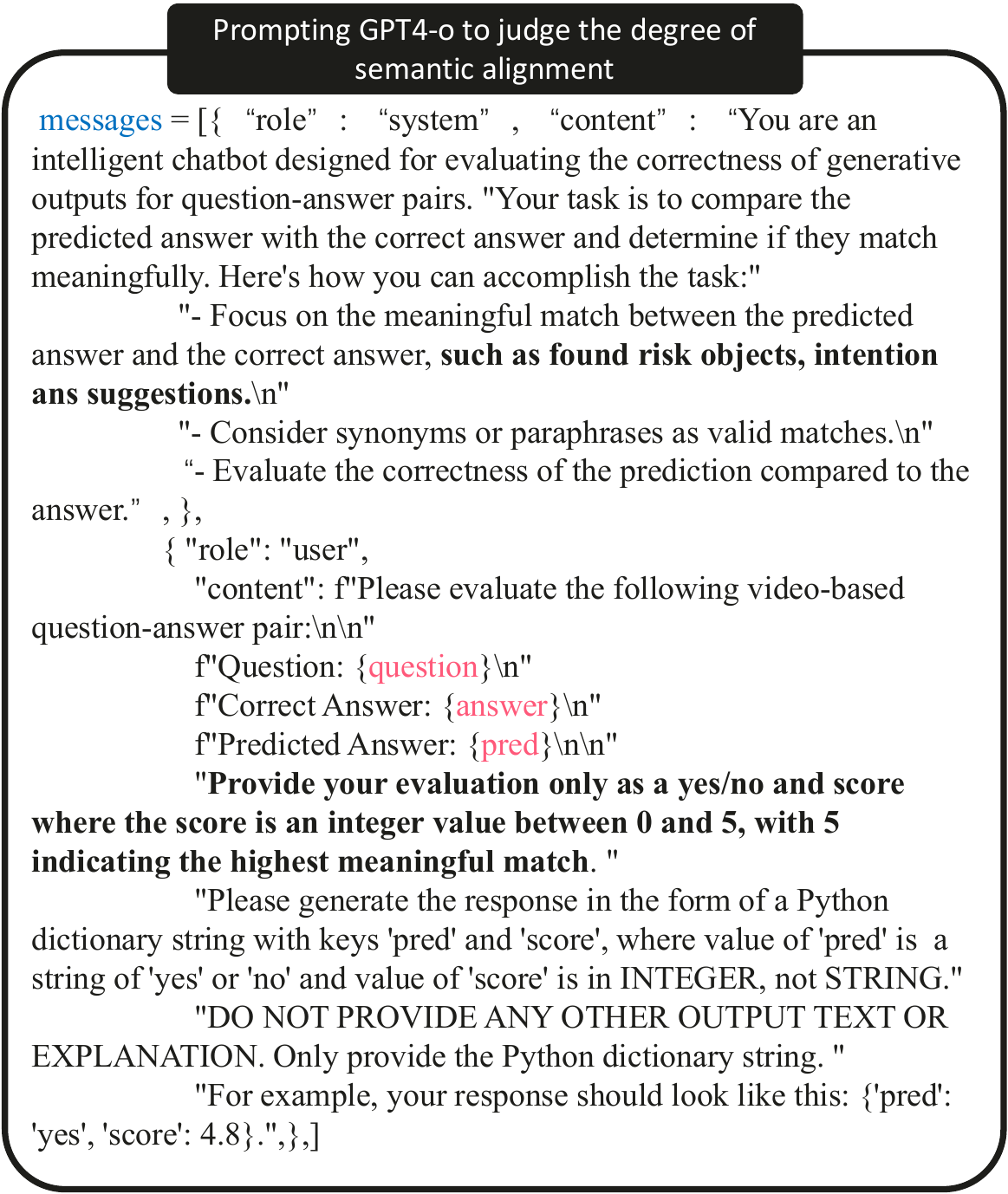}
    \caption{\rev{\textbf{Prompts we use to prompt GPT-4o to generate the degree of semantic alignment between the predictions and the ground-truths.}}
    }
    \label{fig:gpt4o_prompt}
    \vspace{-3mm}
\end{figure}

\noindent \textbf{Evaluation metrics.}
ROLISP comprises two tasks: (1) captioning to identify and explain risk objects while predicting ego-vehicle intentions and motions, and (2) risk object detection.
The captioning performance follows standard metrics~\cite{malla2023drama},~\ie, BLEU-4 (B4), METEOR (M), CIDER (C), and SPICE (S).
\rev{We also utilize the LLM, ~\ie, GPT-4o~\cite{gpt4o}, to assess the degree of semantic alignment between the predicted results and the ground truth. The degree of semantic alignment is rated on a scale from 0 to 5, with 0 representing the lowest and 5 representing the highest score. The final result is the average score of all the predicted samples. The detailed prompts for GPT-4o are shown in Fig.~\ref{fig:gpt4o_prompt}.
}
We employ the mean intersection over union (mIoU) for detection assessment. Additionally, we present IoU scores categorized by object size: small (IoU$_{S}$), medium (IoU$_{M}$), and large (IoU$_{L}$).
%

%--------------------------------
%------------------------------------

%-----------------------------
\begin{figure*}[t]
    \centering
    \includegraphics[width=\linewidth]{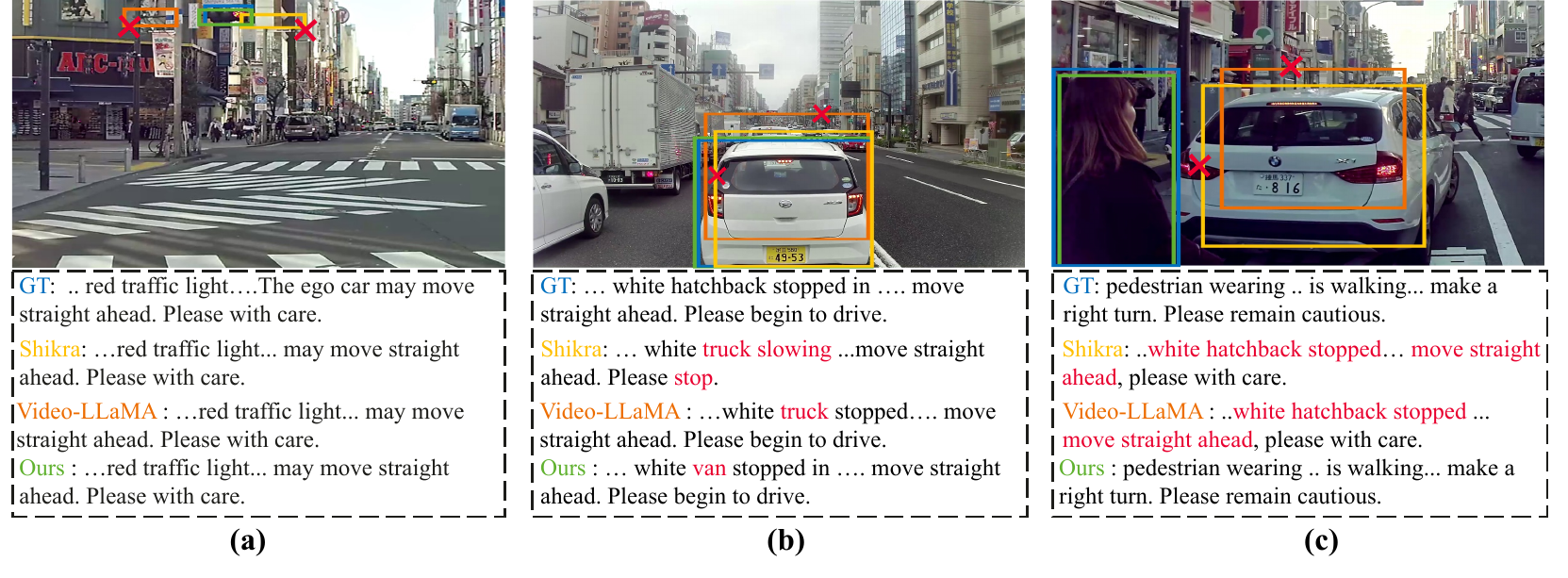}
    % \vspace{-2.5mm}
    \vspace{-2.0em}
    \caption{\textbf{The visualization comparison with the the state-of-the-art.} For clarity, we only show some keywords of the full captions. The inaccurate bounding box and error-predicted words in captions are marked by \textcolor{red}{red}. Compared with the current MLLMs, Our model can \textbf{(a)} detect the small object,~\ie, traffic light, \textbf{(b)} generate more precise bounding boxes and \textbf{(c)} accurately identify the highest risk object.
    }
    \label{fig:visualize}
    \vspace{-3mm}
\end{figure*}
%------------------------------

%-----------------------------
\begin{figure*}[t]
    \centering
    \includegraphics[width=0.99\linewidth,height=0.18\textheight]{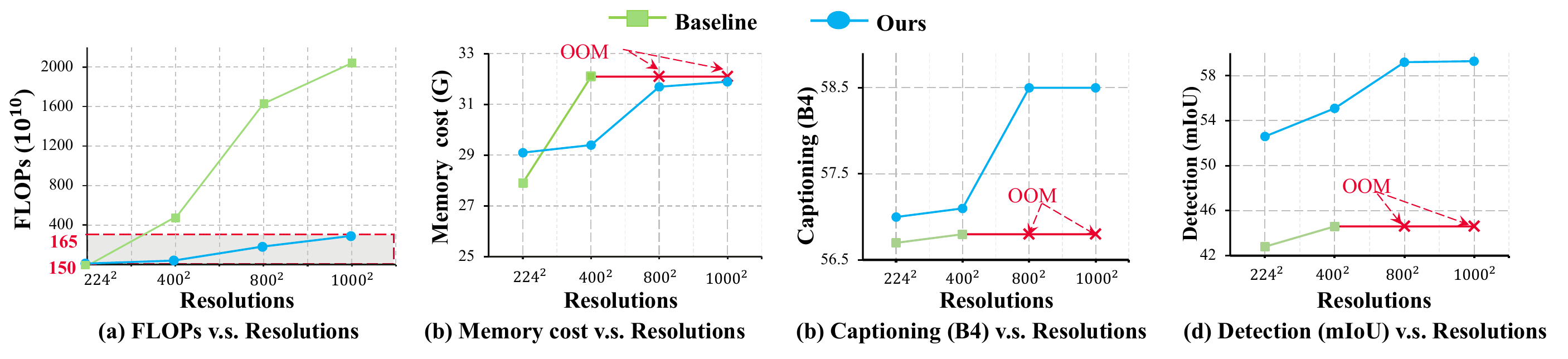}
    \caption{\textbf{Comparison of the baseline and ours across different resolution inputs.} 
    The baseline processes varying resolution videos, while our method, thanks to the spatial perception stream, uses a single high-resolution image. "OOM" indicates out of memory. FLOPs and memory metrics are based on a V100 GPU with a batch size of one, and FLOPs values are scaled to $10^{10}$ for clarity. 
\textbf{(a)} FLOPs v.s. Resolutions. \textbf{(b)} Memory cost v.s. Resolutions. \textbf{(c)} Captioning (B4) v.s. Resolutions. \textbf{(d)} Detection (mIoU) v.s. Resolutions.
    }
    \label{fig:compare}
    \vspace{-3mm}
\end{figure*}
%------------------------------

 % \rowcolor{mygray}
%------------------------------------
\subsection{Comparison with the State-of-the-Art Methods}
% In order to demonstrate the effectiveness of our proposed HiLM-D,
We conduct the experiments on DRAMA-ROLISP to compare with both image-based and video-based MLLMs, including BLIP-2~\cite{li2023blip}, MiniGPT-4~\cite{zhu2023minigpt}, LLaVA-1.5~\cite{liu2023improvedllava}, InstrutBLIP~\cite{Dai2023instruct}, Shikra~\cite{chen2023shikra}, Monkey,~\cite{li2023monkey}, \rev{Qwen-VL~\cite{Qwen-VL}, LLaVA-Next~\cite{liu2024llavanext}} eP-ALM~\cite{shukor2023ep}, and Video-LLaMA~\cite{zhang2023video}, \rev{Video-LLaMA 2~\cite{damonlpsg2024videollama2}}; 
\rev{For all models, we select the 7B parameter version of the LLM.}
Note that except for Shikra and \rev{Qwen-VL}, other models have no ability to detect objects.
Hence, we integrate the query-aware detector into them for bounding boxes,~\ie, replacing $\mathbf{I}_i$ to $\mathbf{Z}_v$ in Eq.~\ref{e:query}, which can be formulated as $ \overline{\mathbf{H}}_i = \text{Cross-Attn}(\mathbf{H},\mathbf{Z}_v)$.

Table~\ref{tab:sota} demonstrates the superior performance of our method in both captioning and detection tasks. Video-based approaches generally surpass image-based ones in captioning, benefiting from temporal data for enhanced risk identification and intention/suggestion prediction. However, their detection capabilities tend to decline, possibly due to the redundant noise in videos affecting object detection. Uniquely, our video-based approach enhances captioning performance (\eg, B4 score increases from $56.3$ to $58.3$) without compromising detection accuracy. Notably, we observe significant improvements for small objects, underscoring the advantages of high-resolution input. \rev{Besides the standard caption metrics, our method also improves the performance of baselines on semantic correctness based on the LLM.}
Fig.\ref{fig:visualize} offers a visual comparison of captioning and detection results from Shikra\cite{chen2023shikra}, Video-LLaMA~\cite{zhang2023video}, and our method, further highlighting our superior performance in both domains.

Except for the MLLMs, we also compare our method with some specialist SOTAs,~\ie, ADAPT~\cite{jin2023adapt} and Grounding-Dino~\cite{liu2023grounding}.
ADAPT is designed for captioning tasks in autonomous driving and Grounding-Dino is a model for language-grounding detection tasks.
Since Grounding-Dino can not generate descriptions for risk objects, we directly use the \textit{`The most risk object'} as prompts, and select the object with the highest confidence score for the detection result.
% From the table, we find our generalized model shows compared results with specialist methods.
Our approach demonstrates comparable performance to existing specialized models. Notably, while Grounding-DINO performs highly in detecting large objects, its efficacy is significantly reduced for small objects. The primary reason for this is Grounding-DINO's inability to accurately identify risk objects, showing a tendency to classify larger objects as risk objects.

\begin{table}
\centering
\caption{\textbf{Ablation study of modules in the spatial perception stream}. `HRES' and \rev{QDA} mean the high-resolution spatial encoder and the query-aware detector respectively. \rev{P indicates the \adapter}. `w/o' indicates our model without the specific module.
}
\begin{footnotesize}
  % \begin{adjustbox}{max width=\linewidth}
  \begin{tabular}{ c| c |c|c|c}
			% \toprule
    % \Xhline{1.0pt}
   % \hhline{----------}
   \Xhline{1.0pt}
    % \toprule
  &  \multirow{2}{*}{Model}  & \multicolumn{1}{c|}{\footnotesize \it Captioning} &\multicolumn{1}{c|}{\footnotesize \it Detection} & \multirow{2}{*}{AVG} \\
    % \cmidrule{4-6}
   & & B4  & mIoU & \\
     % \cmidrule{1-8}
     \hline
     % \midrule
%    & \\
% \transparent{0.5}$(a)$ & \transparent{0.5}Ours & \transparent{0.5}{55.8} & \transparent{0.5}{59.6} & \transparent{0.5}{57.7} \\
\rowcolor{mygray}$(a)$ & Ours & \textbf{56.3} & \textbf{60.5} & \textbf{58.4} \\
$(b)$ &  w/o HRES & 55.8 & 43.7~\subtraction{-16.8} & 49.8 \\
$(c)$ &   w/o \rev{QDA} & 56.1 & 47.5~\subtraction{-13.0}  & 51.8 \\
$(d)$ &  w/o P & 53.1~\subtraction{-3.2}  & 53.8~\subtraction{-6.7} & 53.5  \\
\hdashline
$(e)$ & Baseline & 50.5~\subtraction{-4.0}&  46.3~\subtraction{-19.1} & 48.4\\
  \hline
	\end{tabular}
    % \end{adjustbox}
\label{tab:eachmodule}
\end{footnotesize}
\end{table}

\subsection{Ablation Study}

In this section, we conduct the ablation study to evaluate and analyze the spatial perception stream (illustrated in Section~\ref{Sec:spatialbranch}), the key part of the proposed \method.
Unless otherwise specified, the base MLLM model used here is MiniGPT-4~\cite{zhu2023minigpt} excluding the ST-Adapter to reduce training overhead.

\definecolor{mygreen}{rgb}{0.9,1,0.9}
\colorlet{lightgreenT}{mygreen!90}

\vspace{0.8mm}
\noindent\textbf{The effect of each module in the spatial perception stream.}
Table~\ref{tab:eachmodule} reports the effect of our proposed modules in the spatial perception stream (SPS).
From the table, we have the following observations:
(i) Compared with the lines $a$ and $f$, we find that the improvements mainly come from our proposed SPS.
(ii) Compared with the line $(a)$-$(c)$, high-resolution spatial encoder (HRES) and query-aware detector (QAD) are essential for the detection performance,~\eg, without HRES and QAD dropping $16.8\%$ and $13.0\%$ respectively in terms of the mIoU score.
(iii) Compared with the line $(a)$, $(d)$, \adapter~benefits both caption and detection performance.
Specifically, the model without \adapter~would degrade $3.2\%$ and $6.7\%$ in terms of B4 and mIoU, since the captured highlighted can not be integrated into the MLLM.

\vspace{1.5mm}
\noindent\textbf{Performance comparison of the baseline and ours across different resolution inputs.}
An intuitive way to enhance the perception ability is to increase the resolution of inputs. 
To further study the effectiveness of our proposed spatial perception stream, we conduct experiments to compare the baseline and our method across different resolution inputs in Fig.~\ref{fig:compare}.
Specifically, inputs of varying resolutions are fed into the baseline and our method respectively.
Note that the baseline model in this ablation is the model with ST-Adapter, hence can receive HR videos, while ours only feed one current HR frame.
%
% we report the FLOPs, memory cost and performance under different input resolutions for each model.
%
From Fig.~\ref{fig:compare},
we can see that as the input video resolution increases, the FLOPs of the baseline model grow proportionally.
For example, as the input resolution escalates from $224^2$ to $1000^2$, the FLOPs of the baseline increase from $156.9$ to $2040.2$, a $13$ times growth.
Differently, benefiting from spatial perception stream, our approach requires only a single high-resolution image, significantly reducing FLOPs, only increasing $1.05$ times.
Furthermore, the baseline model encounters OOM issues when the resolution exceeds $400^2$. In contrast, our approach remains stable up to a resolution of $1000^2$.
In summary, our method can outperform the baseline model while using much less computation and memory cost.

\begin{table}
\centering
\caption{\textbf{Effect of different caption generators}. We compare the performance of different caption generators to evaluate the effect of LLM. Our default setting is marked in \colorbox{mygray}{gray}.
}
\begin{footnotesize}
  % \begin{adjustbox}{max width=\linewidth}
  \begin{tabular}{ l | l |c|c}
			% \toprule
    \Xhline{1.0pt}
   & \multirow{2}{*}{Generator}  & \multicolumn{1}{c|}{\footnotesize \it Captioning} &\multicolumn{1}{c}{\footnotesize \it Detection}  \\
    % \cmidrule{4-6}
  &  & B4  & mIoU \\
     % \cmidrule{1-8}
     \hline 
   \multirow{2}{*}{LLM} &  OPT-1.3B~\cite{zhang2022opt} & 54.8 & 58.3  \\
    & \cellcolor{mygray}{Vicuna-7B~\cite{chiang2023vicuna}} & \cellcolor{mygray}{\textbf{56.3}} & \cellcolor{mygray}{\textbf{60.5}}\\
     \hdashline 
     \multirow{2}{*}{Decoder} & ADAPT~\cite{jin2023adapt} & 52.6 & 54.9 \\
      &  DRAMA~\cite{malla2023drama} & 51.3 & 53.8  \\
%    & \\
  \hline
	\end{tabular}
    % \end{adjustbox}
\label{tab:generator}
\end{footnotesize}
% \vspace{-3mm}
\end{table}

\begin{table}
\centering
\caption{\textbf{The versatility of spatial perception stream (SPS)}. We evaluate both image and video-based methods, specifically BLIP-2 and Video-LLaMA, under consistent settings.
}
\begin{footnotesize}
  % \begin{adjustbox}{max width=\linewidth}
  \begin{tabular}{  l |c|c|c}
			% \toprule
    \Xhline{1.0pt}
    \multirow{2}{*}{Model}  & \multicolumn{1}{c|}{\footnotesize \it Captioning} &\multicolumn{1}{c|}{\footnotesize \it Detection} & \multirow{2}{*}{AVG} \\
    % \cmidrule{4-6}
    & B4  & mIoU & \\
     % \cmidrule{1-8}
     \hline
    BLIP-2~\cite{li2023blip} & 46.1 & 46.3 & 46.2 \\
     + SPS &  50.9 & 59.8 & 55.4  \\
     above & \addition{+4.8} & \addition{+13.5} &  \addition{+9.2} \\
     \hdashline 
     Video-LLaMA~\cite{zhang2023video} & 53.9 & 42.8 & 48.4\\
     + SPS & 55.9& 59.3 & 57.6 \\
     above & ~\addition{+2.0}  &~\addition{+16.5} & \addition{+9.2} \\
     
%    & \\

  \hline
	\end{tabular}
    % \end{adjustbox}
\label{tab:applied}
\end{footnotesize}
% \vspace{-3mm}
\end{table}

%-------------
\begin{table}[t]
\centering
\caption{\textbf{Ablation study of the frame selected from the video}. The model performs best when selecting the last frame.
}
\begin{footnotesize}
  % \begin{adjustbox}{max width=\linewidth}
  \begin{tabular}{  l |c|c|c}
			% \toprule
    \Xhline{1.0pt}
    \multirow{2}{*}{Frame}  & \multicolumn{1}{c|}{\footnotesize \it Captioning} &\multicolumn{1}{c|}{\footnotesize \it Detection} & \multirow{2}{*}{AVG} \\
    % \cmidrule{4-6}
    & B4  & mIoU & \\
     % \cmidrule{1-8}
     \hline
    first & 52.5 & 50.8 & 54.0 \\
    middle  & 55.6 & 57.1 &  56.4 \\
  \rowcolor{mygray}  last  &   \textbf{56.3} &  \textbf{60.5} & \textbf{58.4}\\
%    & \\
  \hline
	\end{tabular}
    % \end{adjustbox}
\label{tab:frame}
\end{footnotesize}
\vspace{-1mm}
\end{table}
%-------------

\vspace{0.8mm}
\noindent\textbf{The effect of LLM.}
To prove the effect of LLM for processing ROLISP, we compare two LLMs,~\ie OPT-1.3B~\cite{zhang2022opt} and Vicuna-7B~\cite{chiang2023vicuna} with two text decoders specific-deigned for \ad,~\ie, ADAPT~\cite{jin2023adapt} and  DRAMA~\cite{malla2023drama}.
Table~\ref{tab:generator} shows that LLM outperforms the general text decoder with a clear margin.
For example, using Vicuna-7B~\cite{chiang2023vicuna}, the model achieves $56.3\%$ in terms of B4, outperforming ADAPT~\cite{jin2023adapt} and DRAMA~\cite{malla2023drama} over $3.7\%$ and $5.0\%$ respectively.
We can also observe that more powerful LLM would benefit the ROLISP,~\ie, Vicuna-7B~\cite{chiang2023vicuna} outperforms OPT-1.3B~\cite{zhang2022opt} over $1.5\%$ and $2.2\%$ in terms of B4 and mIoU.

\vspace{0.8mm}
\noindent\textbf{The versatility of spatial perception stream.}
Table~\ref{tab:applied} demonstrates that our proposed spatial perception stream can be applied to existing MLLMs to improve their performance on ROLISP.
Specifically, with our spatial perception stream, BLIP-2~\cite{li2023blip} sees improvements of $4.8\%$ in B4 and $13.5\%$ in mIoU scores, while Video-LLaMA~\cite{zhang2023video} benefits with gains of $2.0\%$ and $16.5\%$ in captioning and detection results, respectively.
% Specifically, combined with our spatial perception stream, BLIP-2~\cite{li2023blip} achieves $4.5\%$ and $13.8\%$ improvements on B4 and mIoU scores respectively.
% %
% Similarly, spatial perception stream can also promote Video-LLaMA~\cite{zhang2023video} over $2.8\%$ and $16.1\%$ on the caption and detection results.

\vspace{0.8mm}
\noindent\textbf{Effect of the sampled frame.}
Table~\ref{tab:frame} analyzes the effect of frames selected at different positions in the spatial perception stream.
`first', 'middle' and `last' mean sampling the first, the middle and the last frames of video as the high-resolution frame fed into the spatial perception stream.
We observe that sampling the last frame outperforms other sampling strategies clearly,~\eg, $4.4\%$ and $2.0\%$ over the first and the middle frames respectively.
The reason is that the assessment of risk objects largely depends on the position of objects in the last frame.

\rev{Furthermore, To address the concern of potentially missing dynamic objects due to sampling a single frame, we conduct additional experiments on our proposed dataset to explore this issue further. Specifically, we considered alternative frame sampling strategies to evaluate whether more frames could help capture sudden changes in object status.
In these additional experiments, we compared four different frame sampling strategies:}

\rev{\begin{itemize}
    \item \textbf{One frame} (the last frame, as used in the original setting),
    \item \textbf{Two frames} (uniformly sampling two frames from the last 1/3 of the video),
    \item \textbf{Three frames} (uniformly sampling three frames from the last 1/3 of the video).
\end{itemize}
}

\rev{Formally, denote the sampled frames as \( \{ \mathbf{F}_{t_i} \}_{i=1}^N \), where \( N \in \{2, 3\} \) is the number of frames sampled from the last 1/3 of the video. These frames are then processed by the high-resolution stream encoder (HRSE) to extract individual feature vectors \( \{ \mathbf{V}_{t_i} \}_{i=1}^N \). The features are then combined using average pooling to produce a unified feature vector \( \overline{\mathbf{V}} \), which has the same dimensionality as the feature vector from a single frame.
This pooled feature vector is then used in subsequent stages, which are consistent with the settings used in the original experiments.
The experimental setup and training/inference configurations were kept identical for all sampling strategies to ensure a fair comparison.}

\rev{From the results in Table~\ref{tab:combiningframe}, we observe that the performance difference between using one or multiple frames is negligible. This indicates that, in most cases of ROLISP, the last frame already contains sufficient information for risk object detection and captioning tasks, as the temporal reasoning stream has already captured the full temporal semantics. Hence, additional frames do not provide significant additional information for these tasks.}

\begin{table}[t]
\centering
\caption{\rev{\textbf{Ablation study of the numbers of sampled frames in the spatial perception branch}. The model performs best when selecting the last frame.}}
\begin{footnotesize}
  \begin{tabular}{  l |c|c|c}
    \Xhline{1.0pt}
    \multirow{2}{*}{\rev{Frame}}  & \multicolumn{1}{c|}{\footnotesize \it \rev{Captioning}} & \multicolumn{1}{c|}{\footnotesize \it \rev{Detection}} & \multirow{2}{*}{\rev{AVG}} \\
    & \rev{B4}  & \rev{mIoU} & \\
     \hline
  \rowcolor{mygray}  \rev{One}  & \rev{56.3} & \rev{60.5} & \rev{58.4} \\
  \rev{Two}  & \rev{56.3} & \rev{60.6} & \rev{58.5} \\
  \rev{Three} & \rev{56.1} & \rev{60.9} & \rev{58.5} \\
  \hline
  \end{tabular}
\label{tab:combiningframe}
\end{footnotesize}
\end{table}

%-------------
\begin{table}[t]
\centering
\caption{\rev{\textbf{Ablation study of joint training of detection and caption tasks}. `joint' and `independent' indicate joint and independent training respectively.
}}
\begin{footnotesize}
  % \begin{adjustbox}{max width=\linewidth}
  \begin{tabular}{  l |c|c|c}
    \Xhline{1.0pt}
     & \multicolumn{1}{c|}{\footnotesize \it \rev{Captioning}} &\multicolumn{1}{c|}{\footnotesize \it \rev{Detection}} & \multirow{2}{*}{\rev{AVG}} \\
    % \cmidrule{4-6}
    & \rev{B4}  & \rev{mIoU} & \\
     % \cmidrule{1-8}
     \hline
   \rev{independent} & \rev{54.3} & \rev{52.5} & \rev{53.4} \\
    \rowcolor{mygray} \rev{joint}  & \rev{\textbf{56.3}} & \rev{\textbf{60.5}} & \rev{\textbf{58.4}} \\
%    & \\
  \hline
  \end{tabular}
  % \end{adjustbox}
\label{tab:joint}
\end{footnotesize}
\vspace{-5mm}
\end{table}
%-------------

\vspace{0.8mm}
\noindent\rev{\textbf{Effect of joint training on detection and captioning.} Table~\ref{tab:joint} demonstrates that training detection and captioning tasks independently leads to a clear decline in performance for both tasks. In contrast, joint training significantly enhances the results across both metrics, indicating a strong mutual benefit.
One possible explanation is that improved object detection enhances risk identification, which in turn contributes to more accurate intent recognition and suggestion generation in the captioning task. Conversely, better captioning performance implies a more effective identification of risk-related objects, thereby boosting detection accuracy. The substantial improvements in both B4 and mIoU scores (from 54.3 to 56.3 and from 52.5 to 60.5, respectively) further validate the advantage of joint training over independent training.
}

%------------------
\begin{table}[t]
\centering
\caption{\textbf{Ablation study on the choice of the high-resolution spatial encoder}. We choose four advanced models where two contain vision-specific prior,~\ie, ResNet50 and ResNet101 and others without the prior,~\ie, ViT-B and ViT-L.
`Pre-train' indicates that we use the weight pre-trained on ImageNet. Note that FLOPs and memory cost are computed for the whole model.
% \vspace{-1.0em}
}
\begin{footnotesize}
  % \begin{adjustbox}{max width=\linewidth}
  \begin{tabular}{  l |c | c |c |c}
			% \toprule
    \Xhline{1.0pt}
    {Model} & {Pretrain} &   {Memory$\downarrow$} &{FLOPs$\downarrow$}   & mIoU$\uparrow$\\
    % \cmidrule{4-6}
     % \cmidrule(lr){1-2}
     \hline
     ResNet50 & \XSolidBrush & \multirow{2}{*}{27.5G} & \multirow{2}{*}{157.5} & 54.0 \\
     ResNet50 & \Checkmark & &  & 55.1 \\
      \hdashline 
     ResNet101 & \XSolidBrush&  \multirow{2}{*}{27.9G}& \multirow{2}{*}{158.6} & 54.5 \\
      ResNet101 & \Checkmark &  &   & 55.9 \\
     \hline 
    ViT-B & \XSolidBrush& \multirow{2}{*}{29.6G} &\multirow{2}{*}{161.4} & 44.7\\
    ViT-B  & \Checkmark& & & 48.9\\
      \hdashline 
    ViT-L & \XSolidBrush& \multirow{2}{*}{31.2G} & \multirow{2}{*}{173.9}  & 45.8\\
     ViT-L & \Checkmark & &   & 49.1 \\
%    & \\
  \hline
	\end{tabular}
    % \end{adjustbox}
\label{tab:HRES}
\end{footnotesize}
\end{table}
\subsection{Effect of Different Designs}
% \vspace{-1.5mm}
Here, we conduct the experiment to analyze the effect of different designs,~\ie, the high-resolution spatial encoder, the \adapter~and the query-aware detector.

\vspace{1.5mm}
\noindent\textbf{High-resolution Spatial Encoder (HRES).}

\vspace{1mm}
\emph{1) Ablation of different backbones:}
% \noindent\textbf{Ablation of different backbones.}
%
In Table~\ref{tab:HRES}, we conduct experiments to analyze the effect of different backbones of HRES. 
To save resources, we use the resolution of $400^2$ to train all models in the same setting.
From the table, we can find that
(1) the plain ViT without the vision-specific prior achieves limited performance on detection;
(2) using the advanced backbones would benefit the detection performance while bringing more memory and computation costs; (3) pre-trained weight would improve the performance.
Since the performance gain obtained by the advanced backbones is limited, we use ResNet50 as the backbone for efficiency.

%-------------
\begin{table}
\centering
\caption{\textbf{Ablation study of multi-scale features $\{ \mathbf{I}_i \}_{i=1}^3$}. The model performs best when using all scales.
}
\begin{footnotesize}
  % \begin{adjustbox}{max width=\linewidth}
  \begin{tabular}{  l |c|c c cc}
			% \toprule
    \Xhline{1.0pt}
    \multirow{2}{*}{Scale}  & \multicolumn{1}{c|}{\footnotesize \it Captioning} &\multicolumn{4}{c}{\footnotesize \it Detection}  \\
    % \cmidrule{4-6}
    & B4  & mIoU & IoU$_{S}$  & IoU$_{M}$  &IoU$_{L}$    \\
     % \cmidrule{1-8}
     \hline
    $\mathbf{I}_1$ & 54.3 & 56.5 & 26.7 & 60.8 & 81.3 \\
     $ \mathbf{I}_1, \mathbf{I}_2$ & 55.7 & 58.5 &  28.4 & 63.8 & 81.1 \\
  \rowcolor{mygray}  $ \mathbf{I}_1, \mathbf{I}_1,  \mathbf{I}_3 $  &  \textbf{56.3} &  \textbf{60.5}& \textbf{33.5} &  \textbf{64.1}& \textbf{81.8} \\
%    & \\
  \hline
	\end{tabular}
    % \end{adjustbox}
\label{tab:multiscale}
\end{footnotesize}
\vspace{-3mm}
\end{table}
%-------------

\vspace{1mm}
\emph{2) Effect of multi-scale features:}
We conduct an ablation study in Table~\ref{tab:multiscale} to evaluate the impact of multi-scale features, \ie, ${ \mathbf{I}_i }{i=1}^3$ in Section~\ref{Sec:spatialbranch}. The results demonstrate that incorporating multi-scale features enhances the detection performance of our model, particularly for medium and small objects. For example, utilizing all scales of features, the model achieves $33.5\%$ in IoU$_s$, surpassing the single-scale model by $6.8\%$. Furthermore, the additional scale information enables the model to perceive objects of various sizes more effectively, leading to more accurate risk object identification.
%-------------
\begin{table}[t]
\centering
\caption{\textbf{Ablation study of the number of the \adapter}. The model performs best when the number is set to $3$.
}
\begin{footnotesize}
  % \begin{adjustbox}{max width=\linewidth}
  \begin{tabular}{  l |c|c|c}
			% \toprule
    \Xhline{1.0pt}
    \multirow{2}{*}{Number}  & \multicolumn{1}{c|}{\footnotesize \it Captioning} &\multicolumn{1}{c|}{\footnotesize \it Detection} & \multirow{2}{*}{AVG} \\
    % \cmidrule{4-6}
    & B4  & mIoU & \\
     % \cmidrule{1-8}
     \hline
    1 & 55.4 & 52.6 & 54.0 \\
    2  & 55.6 & 57.1 &  56.4 \\
  \rowcolor{mygray}  3  &   \textbf{56.3} &  \textbf{60.5} & \textbf{58.4}\\
    4  & 56.2 & 60.4 & 58.3\\
%    & \\
  \hline
	\end{tabular}
    % \end{adjustbox}
\label{tab:numberofim}
\vspace{-3mm}
\end{footnotesize}
\end{table}
%-------------

%-------------
\begin{table}[t]
\centering
\caption{\textbf{Ablation study of $\alpha$ in Eq.~\ref{e:xadapter}}. `w/' and `w/o' indicate with and without $\alpha$ respectively.
}
\begin{footnotesize}
  % \begin{adjustbox}{max width=\linewidth}
  \begin{tabular}{  l |c|c|c}
			% \toprule
    \Xhline{1.0pt}
     & \multicolumn{1}{c|}{\footnotesize \it Captioning} &\multicolumn{1}{c|}{\footnotesize \it Detection} & \multirow{2}{*}{AVG} \\
    % \cmidrule{4-6}
    & B4  & mIoU & \\
     % \cmidrule{1-8}
     \hline
    w/o & 54.7 & 56.5 & 55.6 \\
    \rowcolor{mygray} w/  & \textbf{56.3} & \textbf{60.5} & \textbf{58.4} \\
%    & \\
  \hline
	\end{tabular}
    % \end{adjustbox}
\label{tab:alpha}
\end{footnotesize}
\vspace{-2mm}
\end{table}
%-------------

%-------------
\begin{table}[t]
\centering
\caption{\textbf{Ablation study of the types of the query-aware detector.} `LLM' indicates directly using LLM to regress the position~\ie, bounding boxes, using numerical values in natural language similar to Shikra~\cite{chen2023shikra}.
`DETR' means the modified decoder used in DETR~\cite{carion2020end}.
}
\begin{footnotesize}
  % \begin{adjustbox}{max width=\linewidth}
  \begin{tabular}{  l |c|c|c}
			% \toprule
    \Xhline{1.0pt}
    \multirow{2}{*}{Type}  & \multicolumn{1}{c|}{\footnotesize \it Captioning} &\multicolumn{1}{c|}{\footnotesize \it Detection} & \multirow{2}{*}{AVG} \\
    % \cmidrule{4-6}
    & B4  & mIoU & \\
     % \cmidrule{1-4}
     \hline
   LLM & 55.8  & 48.9& 52.4\\
   DETR~\cite{carion2020end} & 55.8 &  56.3& 56.1\\
  \rowcolor{mygray}  Ours  &  \textbf{56.3} &  \textbf{60.5} & \textbf{58.4}\\
%    & \\
  % \bottomrule
  \hline
	\end{tabular}
    % \end{adjustbox}
\label{tab:qdh}
\end{footnotesize}
% \vspace{-2mm}
\vspace{-3mm}
\end{table}
%-------------
\begin{table}[t]
\centering
\caption{\textbf{Comparing different training types.}
`LoRA' indicates the efficient fine-tuning of the LLM using LoRA~\cite{hu2021lora}.
`Frozen' means freeze the LLM during training.
`Trainable Param' means the number of trainable parameters.
}
\begin{footnotesize}
  % \begin{adjustbox}{max width=\linewidth}
  \begin{tabular}{  c | c | c|c}
			% \toprule
    \Xhline{1.0pt}
    \multirow{2}{*}{Type} & \multirow{2}{*}{Trainable Param} & \multicolumn{1}{c|}{\footnotesize \it Captioning} &\multicolumn{1}{c}{\footnotesize \it Detection}  \\
    % \cmidrule{4-6}
   &   & B4  & mIoU \\
     % \cmidrule{1-4}
     \hline
   LoRA & 115.7M & 56.1 &  59.8\\
  \rowcolor{mygray}  Frozen  & \textbf{106.8M} & \textbf{56.3} &  \textbf{60.5} \\
%    & \\
  % \bottomrule
  \hline
	\end{tabular}
    % \end{adjustbox}
\label{tab:training}
\end{footnotesize}
% \vspace{-3mm}
\end{table}
%-------------

\vspace{1.5mm}
\noindent\textbf{\adapter.}

%-------------

\vspace{1mm}
\emph{1) The number of \adapter:}
% \subsubsection{The number of \adapter}
%
In Table~\ref{tab:numberofim}, we study the effect of numbers of \adapter.
We uniformly incorporate \adapter~into the different layers of the CLIP-ViT.
We find that the model accuracy saturates when the numbers go larger.
For example, as the number arises from $1$ to $3$, the mIoU achieves $7.9\%$ improvement.
We also find that adding more \adapter~cannot monotonically promote performance when the number is larger than $3$. Therefore, we empirically set the number to $3$ by default.

\vspace{1mm}
\emph{2) Effect of $\alpha$:}
% \subsubsection{The number of \adapter}
%
In Table~\ref{tab:alpha}, we conduct experiments to prove the effect of $\alpha$ in Eq.~\ref{e:xadapter}.
Specifically, using $\alpha$ would achieve $1.6\%$ and $4.0\%$ improvement in terms of B4 and mIoU, compared with without $\alpha$.

%--------------------------------
\begin{figure}[t]
    \centering
    \includegraphics[width=\linewidth,height=0.1\textheight]{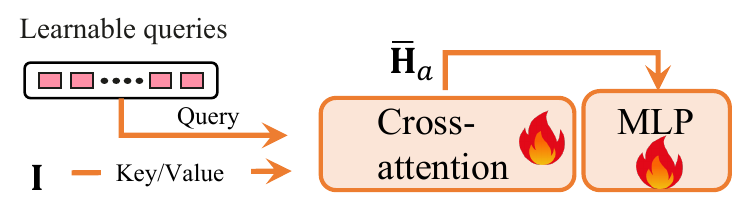}
    \caption{\textbf{DETR-like detector.} Using learnable queries to capture semantics from $\mathbf{I}$ for detection. 
    }
    \label{fig:detector}
\end{figure}
%------------------------------------

\vspace{1.5mm}
\noindent\textbf{Types of the query-aware detector (QAD).}
Table~\ref{tab:qdh} shows the experiments to analyze the effect of the different types of QAD.
Besides ours shown in Fig.~\ref{fig:architecture} (b), we select two other modules to compare,~\ie, directly using LLM to regress the bounding boxes using numerical values as Shikra~\cite{chen2023shikra} and the modified decoder in DETR~\cite{carion2020end}, which is detailed in the following:
%
 % we illustrate the detailed architectures of different types except ours,~\ie, LLM and DETR. 
%

\vspace{1.5mm}
\noindent\textbf{LLM.}
Specifically, `LLM' indicates that directly using LLM to regress
the bounding boxes as Shikra~\cite{chen2023shikra}.
To this end, we integrate the positions of bounding boxes into the captions as this: 
\begin{tcolorbox}
{The location is at $[x1, y1, x2, y2 ]$},
\end{tcolorbox}
where $[x1, y1, x2, y2 ]$ indicates the top-left and bottom-right of the bounding box.
For example, given the full caption:
\begin{tcolorbox}
There is a pedestrian wearing black pants and a white shirt, crossing to the left side of the street, in front of the ego car. The ego car intends to proceed straight. Please kindly give way.
\end{tcolorbox}
And the corresponding bounding box for the pedestrian is $[1264, 756, 1324, 939]$.
We first normalize the coordinate by $\left[1264/W, 756/H, 1324/W, 939/H \right ]$, where $H$ and $W$ is the height and width of the image (in this case $H=1520$, $W=2704$).
Hence, the normalized bounding box is $\left[  0.465, 0.495, 0.490, 0.614 \right]$.
Then, the final caption can be:
\begin{tcolorbox}
There is a pedestrian wearing black pants and a white shirt, crossing to the left side of the street, in front of the ego car. The ego car intends to proceed straight. Please kindly give way. The Location is at $\left[  0.465, 0.495, 0.490, 0.614 \right]$.
\end{tcolorbox}

\vspace{1.5mm}
\noindent\textbf{DETR.} The implementation of the QAD in the DETR~\cite{carion2020end} manner,~\ie, using the learned query embedding, is shown in Fig.~\ref{fig:detector}.

The results show that using LLM achieves limited performance in detection, which shows that our method is more effective in this small-scale detection dataset.
%
% The reason may be that we do not use the large object detection data to pre-train the LLM similar as Shikra~\cite{chen2023shikra}\hjh{re-write reason, to say our method is more effective in this small-scale detection dataset}. 
%
Moreover, ours performs better than DETR,~\eg, $59.6\%$ vs $56.3\%$ on the detection task due to prior knowledge.

\vspace{1.5mm}
\noindent\textbf{Frozen v.s. LoRA-Finetuning.}
Table~\ref{tab:training} reports comparing the effect of freezing and efficient fine-tuning of the LLM.
We use LoRA~\cite{hu2021lora} to fine-tune LLM efficiently.
The results show freezing the LLM can achieve better performance with more efficiency.

% \vspace{-1mm}

\begin{table}[t]
\centering
\caption{\textbf{Comparing different position representations.} `Vocab.' means to use extra vocabularies to represent coordinates, like~\cite{wang2023visionllm}, and `Numerical' means to directly use numerals in natural language to express coordinates as~\cite{chen2023shikra}.
}
\begin{footnotesize}
  % \begin{adjustbox}{max width=\linewidth}
  \begin{tabular}{  c |c|c|c}
			% \toprule
    \Xhline{1.0pt}
    \multirow{2}{*}{Type}  & \multicolumn{1}{c|}{\footnotesize \it Captioning} &\multicolumn{1}{c|}{\footnotesize \it Detection} & \multirow{2}{*}{AVG} \\
    % \cmidrule{4-6}
    & B4  & mIoU & \\
     % \cmidrule{1-4}
     \hline
   Vocab. & 55.8  & 45.6& 50.7\\
   Numerical & 56.3 &  50.5& 53.4\\
  \rowcolor{mygray}  Ours  &  \textbf{56.3} &  \textbf{60.5} & \textbf{58.4}\\
%    & \\
  % \bottomrule
  \hline
	\end{tabular}
    % \end{adjustbox}
\label{tab:position}
\end{footnotesize}
\end{table}
%-------------

%-------------
\begin{table}[t]
\centering
\caption{\textbf{Experiments on the generalized dataset Shikra-RD~\cite{chen2023shikra}.}
Our method can improve the performance of MLLMs in the generalized domain.
`HRS' indicates the spatial perception stream.
% `DETR' means the modified decoder used in DETR~\cite{carion2020end}
}
\begin{footnotesize}
  % \begin{adjustbox}{max width=\linewidth}
  \begin{tabular}{  l |c|c|c}
			% \toprule
    \Xhline{1.0pt}
    \multirow{2}{*}{Method}  & \multicolumn{1}{c|}{\footnotesize \it Captioning} &\multicolumn{1}{c|}{\footnotesize \it Detection} & \multirow{2}{*}{AVG} \\
    % \cmidrule{4-6}
    & B4  & mIoU & \\
     % \cmidrule{1-4}
     \hline
   MiniGPT-4~\cite{zhu2023minigpt} & 26.7 & 47.5& 37.1 \\
   + SPS & 28.1& 53.6 &40.9\\
   \hline
   InstrutBILP~\cite{Dai2023instruct} & 28.3  & 47.8& 38.1\\
   + SPS & \textbf{29.3} &  \textbf{53.4}& \textbf{41.4}\\
%    & \\
  % \bottomrule
  \hline
	\end{tabular}
    % \end{adjustbox}
\label{tab:shikrard}
\end{footnotesize}
\end{table}
%-------------

\vspace{1.5mm}
\noindent\textbf{Different position representations for LLM.}
For detect objects in the autoregressive model, several methods~\cite{wang2023visionllm} introduce extra vocabularies (~\eg, $<bin0>$, · · ·, $<bin1000>$) to represent coordinates for object detection in spatially discretized images. In contrast, Shikra~\cite{chen2023shikra} represents coordinates naturally and intuitively, using numbers directly. Here, we also conduct experiments to compare these two different position representations for LLM, when directly using LLM for producing bounding boxes.
As shown in Table~\ref{tab:position}, using numerical directly achieves better performance compared with adding extra vocabularies.
Also, using extra vocabularies would degrade the captioning performance.
This is because using extra vocabularies requires fine-tuning the LLM, thus impairing the capabilities of the original LLM.

\subsection{\rev{Open-loop Evaluation on Trajectory Prediction}}

\rev{In this section, we also conduct the open-loop evaluation on trajectory prediction to better demonstrate the effectiveness of our approach. Since our proposed method, ROLISP, is primarily designed for language-based trajectory prediction rather than direct trajectory forecasting, we adapt the NuScenes dataset to incorporate the evaluation of these trajectory metrics. Specifically, we constructed a language-based trajectory prediction task from NuScenes [1] to better evaluate the performance of our method.}

\rev{To clarify the task setup, the QA pairs used for trajectory prediction are as follows:
\begin{tcolorbox}
{\textcolor{red}{Question}: What are the future trajectories of the ego-car?}\\
{\textcolor{red}{Answer}: (-5.0, 2.3), (-2.0, 3.2), ..., (10.2, 9.6)}
\end{tcolorbox}
Here, \((x, y)\) represents the coordinates in bird’s-eye view (BEV) for future frames.
We sampled 5 consecutive keyframes from NuScenes, where the first two frames are used as input to predict the motion trajectories for the subsequent three frames. We trained the model on a total of 1,000 videos from the train split of NuScenes and evaluated it on 200 videos from the validation split. For each trajectory prediction, we calculated the ADE (Average Displacement Error) and FDE (Final Displacement Error) to assess the performance of our method. We followed the data processing and evaluation setup of the UniAD [2] for this experiment.}

\rev{The results in Table \ref{tab:open-loop} show that our method improves the baseline models (MiniGPT-4 and Video-LLaMA) with better performance in both ADE and FDE metrics. This improvement demonstrates the fine-grained perception provided by our model.}

\rev{We have not included closed-loop evaluation metrics because our approach does not directly involve trajectory control or real-time feedback mechanisms. The focus of our work is on improving the fine-grained perception capabilities of MLLMs. We plan to explore such settings in future experiments to further enhance the robustness and adaptability of our model.}

\subsection{Apply to Generalized Domain}
% \vspace{-1mm}
To explore the effect of our spaitial perception stream on the generalized domain, we also conduct experiments on Shikra-RD~\cite{chen2023shikra}.
Shikra-RD is a dataset for referential dialogue, which is constructed from Flickr30K Entities~\cite{plummer2015flickr30k} by GPT-4~\cite{OpenAI_2023}.
Specifically, besides responding to queries, the task involves providing bounding boxes for objects mentioned in the answers, which is very similar to our ROLISP.
We select two MLLMs,~\ie, MiniGPT-4~\cite{zhu2023minigpt} and InstructBLIP~\cite{Dai2023instruct}, as the baselines and augment them with our spatial perception stream.
The input resolution for the spatial perception stream is $448 \times 448$ here.
The results of their performance on Shikra-RD are reported in Table~\ref{tab:shikrard}.
Results prove our spatial perception stream can benefit the baselines, especially in the detection performance,~\eg, outperforming InstructBLIP over $3.8\%$.

\begin{table}[t]
\centering
\caption{\rev{\textbf{Open-loop Planning on Sampled QA pairs from NuScenes}. `ADE' indicates the average displacement error and `FDE' means the final displacement error.}}
  \begin{tabular}{  l |c | c}
    \Xhline{1.0pt}
    \rev{Method}  & \rev{ADE} & \rev{FDE}\\
     \hline
  \rev{MiniGPT-4} & \rev{2.42} & \rev{5.15}\\
   \rev{MiniGPT-4 + Ours} & \rev{2.18} & \rev{4.02} \\
   \rev{Video-LLaMA} & \rev{2.39} & \rev{4.81} \\
   \rev{Video-LLaMA + Ours} & \rev{1.98} & \rev{3.66} \\
  \hline
  \end{tabular}
\label{tab:open-loop}
\end{table}

\section{Conclusion and Limitations}

In this paper, we focus on Risk Object Localization and Intention and Suggestion Prediction (ROLISP), which involves identifying the most important traffic objects and their bounding boxes, providing explanations, and predicting the next intention for the ego car.
Considering the limited visual perception ability of pre-trained CLIP-ViT in existing multimodal large language models (MLLMs), we introduce \method, a two-stream framework designed to enhance visual information processing for ROLISP. Our framework captures comprehensive visual information, including temporal, multi-scale, and high-resolution details.
The temporal reasoning stream uses a static visual encoder with trainable, lightweight ST-Adapters, enabling CLIP-ViT to effectively process dynamic video content. The spatial perception stream features a high-resolution spatial encoder and an \adapter, which captures multi-scale information and integrates it into the temporal stream, ensuring complete information for ROLISP.
Notably, our spatial perception stream is lightweight, training-efficient, and easily integrates into existing MLLMs. Experiments on the ROLISP benchmark show \method's significant advantages over leading MLLMs, with improvements of $3.7\%$ in BLEU-4 for captioning and $8.7\%$ in mIoU for detection.

\vspace{1.0mm}
\noindent \textbf{Limitations.}
In our dataset, each video contains only one risk object, which might not capture the complexity of real-world scenarios. 
The dataset also lacks extreme weather conditions like rain or fog and multi-view information, which are crucial for autonomous driving.
\rev{Furthermore, data samples in our dataset do not contain too many dynamic scenes related to missing fast-moving or sudden objects, especially in the last few frames.}
% Furthermore, the suggestions provided are typically concise, such as 'stop' or 'yield', potentially oversimplifying the range of possible actions.
% %
% % \vspace{-0.2mm}
We will curate a more diverse and challenging dataset to advance the field in the future.
%
% Multi-view and multi-object.Multi-view and multi-object.Multi-view and multi-object.

\noindent\textbf{Data Availability Statement} Our dataset are constructed based on DRAMA: \url{https://usa.honda-ri.com/drama}.

\bibliography{sn-bibliography}% common bib file

\end{document}